\documentclass{article}

\usepackage{microtype}
\usepackage{graphicx}
\usepackage{subfigure}
\usepackage{booktabs} 
\usepackage{amsmath}

\usepackage{hyperref}

\usepackage[accepted]{icml2018}

\usepackage{amsthm}
\usepackage{breqn}
\usepackage[capitalize]{cleveref}
\usepackage{amsfonts,amssymb}
\usepackage{amsopn}

\usepackage{tikz}
\usetikzlibrary{matrix}
\usepackage{framed}
\usepackage{xspace}
\usepackage{bm}

\renewcommand{\algorithmicrequire}{\textbf{Input:}}
\renewcommand{\algorithmicensure}{\textbf{Output:}}
\newcommand{\constraint}{\mathcal{K}}
\newcommand{\tnabla}{\tilde{\nabla}}
\newcommand{\bd}{\mathbf{d}}
\newcommand{\bx}{\mathbf{x}}
\newcommand{\ba}{\mathbf{a}}
\newcommand{\bv}{\mathbf{v}}
\newcommand{\by}{\mathbf{y}}
\newcommand{\bta}{\tilde{\mathbf{a}}}
\newcommand{\expect}{\mathbb{E}}

\newcommand{\regret}{\mathcal{R}} 
\newcommand{\sregret}{\mathcal{SR}} 
\newcommand{\oregret}{\mathcal{R}_T^{\mathcal{E}}} 
\newcommand{\reals}{\mathbb{R}}
\newcommand{\ground}{\Omega}
\newcommand{\setsys}{\mathcal{I}}

\newcommand{\AlgMetaVR}{\textsf{Meta-FW w/ VR}\xspace}
\newcommand{\AlgMetaNVR}{\textsf{Meta-FW w/o VR}\xspace}
\newcommand{\AlgOGA}{\textsf{OGA}\xspace}
\newcommand{\AlgOGD}{\textsf{OGD}\xspace}
\newcommand{\AlgOnlineVR}{\textsf{OS-FW w/ VR}\xspace}
\newcommand{\AlgOnlineNVR}{\textsf{OS-FW w/o NVR}\xspace}
\newcommand{\AlgRegu}{\textsf{Regularized-OFW}\xspace}

\newcommand{\AlgOG}{\textsf{Online Greedy}\xspace}

\newcommand{\AlgMetaFW}{\textsf{Meta-Frank-Wolfe}\xspace}
\newcommand{\AlgOneShotFW}{\textsf{One-Shot Frank-Wolfe}\xspace}

\DeclareMathOperator{\Tr}{Trace}

\DeclareMathOperator*{\argmax}{arg\,max}
\DeclareMathOperator*{\argmin}{arg\,min}
\newcounter{assump}
\newtheorem{theorem}{Theorem}

\theoremstyle{definition}
\newtheorem{assumption}[assump]{Assumption}

\icmltitlerunning{Projection-Free Online Optimization with Stochastic 
Gradient: From Convexity to Submodularity}

\begin{document} 
	\makeatletter
	\def\ICML@appearing{\relax}
	\makeatother
	
	
	\author{\name Lin Chen \email lin.chen@yale.edu \\
		\addr Yale Institute for Network Science\\
		Department of Electrical Engineering\\
		Yale University\\
		New Haven, CT 06511, USA
		\AND
		\name Christopher Harshaw \email christopher.harshaw@yale.edu \\
		\addr Yale Institute for Network Science\\
		Department of Computer Science\\
		Yale University\\
		New Haven, CT 06511, USA
		\AND
	\name Hamed Hassani \email hassani@seas.upenn.edu \\
	\addr Department of Electrical and Systems Engineering\\
	University of Pennsylvania\\
	Philadelphia, PA 19104, USA
	\AND
	\name Amin Karbasi \email amin.karbasi@yale.edu \\
	\addr Yale Institute for Network Science\\
	Department of Electrical Engineering\\
	Yale University\\
	New Haven, CT 06511, USA
}

\twocolumn[
	\icmltitle{Projection-Free Online Optimization with Stochastic 
		Gradient: From Convexity to Submodularity}
	
	
	
	\icmlsetsymbol{equal}{*}
	
	\begin{icmlauthorlist}
		\icmlauthor{Lin Chen}{yins,yaleee}
		\icmlauthor{Christopher Harshaw}{yins,yalecs}
		\icmlauthor{Hamed Hassani}{upenn}
		\icmlauthor{Amin Karbasi}{yins,yaleee}
	\end{icmlauthorlist}
	\icmlaffiliation{yins}{Yale Institute for Network Science, Yale University, 
		New Haven, CT, USA}
	\icmlaffiliation{yaleee}{Department of Electrical Engineering, Yale 
	University
}
	\icmlaffiliation{yalecs}{Department of Computer Science, Yale University
}
	\icmlaffiliation{upenn}{Department of Electrical and Systems Engineering, 
University of Pennsylvania, Philadelphia, 
	PA, USA}
	
	\icmlcorrespondingauthor{Lin Chen}{lin.chen@yale.edu}
	
	\icmlkeywords{Machine Learning, ICML}
	
	\vskip 0.3in
	]
	
	
	
	\printAffiliationsAndNotice{}  

\begin{abstract} 
Online optimization has been a successful framework for solving large-scale problems under 
computational constraints and partial information. 
Current methods for online convex optimization require either a projection or exact gradient 
computation at each step, both of which can be prohibitively expensive for large-scale applications.
At the same time, there is a growing trend of non-convex optimization in machine learning community 
and a need for online methods. Continuous DR-submodular functions, which 
exhibit a natural 
diminishing returns condition, have recently been proposed as a broad class of non-convex functions 
which may be efficiently optimized. Although online methods have been introduced, they suffer from 
similar problems. 
%
%
In this work, we propose \AlgMetaFW, the first online projection-free algorithm that uses stochastic 
gradient 
estimates. The algorithm relies on a careful sampling of gradients in each round and achieves the 
optimal $O(\sqrt{T})$ adversarial regret bounds for convex and continuous submodular 
optimization.
We also propose \AlgOneShotFW, a simpler algorithm which requires only a single stochastic gradient 
estimate in each round and achieves an $O(T^{2/3})$ stochastic regret bound for 
convex and 
continuous 
 submodular optimization.
We apply our methods to develop a novel ``lifting'' framework for the online discrete submodular 
maximization and also see that they outperform current 
state-of-the-art techniques on various experiments.
%
\end{abstract} 

\section{Introduction}
\label{sec:intro}
As the amount of collected data becomes massive in both size and complexity, algorithm designers are 
faced with unprecedented challenges in statistics, machine learning, and control. 
In the  past decade, online optimization has provided a successful computational framework for 
tackling a wide variety of challenging problems, ranging from non-parametric regression to portfolio 
management \cite{Calandriello2017Second, Agarwal2006Algorithms}. 
In online optimization, a large or complex optimization problem is broken 
down into a sequence of smaller optimization problems, each of which must be solved with limited
information. This framework captures many real-world scenarios in which standard optimization theory 
does not apply. For instance, a machine learning application cannot feasibly process terabytes of data 
at a single time; rather, subsets of data may be handled in a sequential fashion. Another example is 
when the 
true objective function is the expectation of an unknown distribution of functions, and may only be 
accessible via samples, as is the case for problems in online learning and control theory 
\cite{Xiao2010Dual, Wang2008Fast}.

Online convex optimization, a branch of online optimization that considers sequentially minimizing 
convex functions, has proved particularly useful for statistical and machine learning applications. 
Online convex optimization has enjoyed much success in these areas because most offline machine 
learning techniques utilize the existing theory of convex optimization.
As in the offline setting, gradient 
methods are a popular 
class of algorithms for online convex optimization due to their simplicity; however, 
they require projections onto the constraint set, which involve solving a quadratic program in the 
general case. These projections are infeasible for large scale applications with complicated constraints  
such as matrix completion, network routing problems, and maximum matchings. Online 
projection-free methods have been proposed and are much more efficient
, replacing a projection onto the constraint set with a 
linear optimization over the constraint set at each iteration \cite{hazan2012projection, 
garber2013linearly}. However, these projection-free methods 
require exact gradient computations, which may be prohibitively expensive for even moderately sized 
data sets and 
intractable when a closed form does not exist. Thus, there is a huge need for online convex 
optimization routines that are projection-free and also robust to stochastic gradient estimates.

While convex programs may be efficiently solved (at least in theory), there is a growing number of 
non-convex problems arising in machine learning and statistics.
Notable examples include nonnegative principle component analysis, low-rank matrix recovery, sigmoid 
loss functions for binary classification, and the training of deep neural networks, to name a few. 
Understanding which types of non-convex functions may be efficiently optimized and developing 
techniques for doing so is a pressing research question for both theory and practice. Recently, 
continuous DR-submodular functions have been proposed as a broad class of non-convex functions 
which admit efficient approximate maximization routines, even though exact maximization is 
NP-Hard \cite{bian16guaranteed}. These functions capture many real-life applications, such as optimal 
experiment design, non-definite quadratic programming, coverage and diversity functions, and 
continuous relaxation of discrete submodular functions. Recent works 
\cite{Chen2018Online} have proposed methods for online continuous DR-submodular 
optimization; 
however, these too require either expensive projections or exact gradient computations.

\paragraph{Our contributions} 
In this paper, we present a suite of projection-free algorithms for online optimization that use 
stochastic estimates of the gradient and leverage the averaging 
technique~\cite{mokhtari2017conditional,mokhtari2018stochastic} to reduce 
their variance. This includes
\begin{itemize}
	\item \AlgMetaFW, the first projection-free algorithm for adversarial online optimization which 
	requires only stochastic gradient estimates. The algorithm relies on a careful sampling of gradients 
	 in each round and achieves optimal $O(\sqrt{T})$ regret and $(1-1/e)$-regret bounds for convex 
	 and submodular optimization, 
	respectively.
	\item \AlgOneShotFW, a simpler projection-free algorithm for stochastic online optimization which 
	requires only a single stochastic gradient estimate in each round. This simpler algorithm achieves 
	$O(T^{2/3})$ regret and $(1-1/e)$-regret bounds for the convex and submodular case, respectively.
	\item A novel class of algorithms for online discrete submodular optimization which are based on 
	lifting discrete 
	functions to the continuous domain, applying our methods with an extremely efficient sampling 
	technique, and using rounding schemes to produce a discrete solution.
\end{itemize}
Finally, to demonstrate the effectiveness of our algorithms, we tested their performance on an 
extensive set 
of experiments and measured against common baselines.

\section{Related Work}\label{sec:related}

The Frank-Wolfe algorithm, also known as the conditional gradient 
descent, was originally proposed for the offline setting in~\cite{frank1956algorithm}. 
The 
framework of online convex optimization was introduced by 
\citet{zinkevich2003online}, in which the online projected gradient 
descent was proposed and proved to achieve an $ O(\sqrt{T}) $ regret bound. However, the 
projections required for such an algorithm are too expensive for many large-scale online problems. The 
online conditional gradient descent was the first projection-free online algorithm, originally proposed 
in~\cite{hazan2012projection}. 
An improved conditional gradient algorithm was later designed for smooth
and strongly convex optimization which achieves the optimal $O(\sqrt{T})$ adversarial regret 
bound~\cite{garber2013linearly}. However, both of these algorithms can perform arbitrarily poorly if 
supplied with 
stochastic gradient estimates.
\citet{lafond2015online} proposed an 
online Frank-Wolfe variant for the any-time stochastic online setting that converges to a stationary 
point for 
non-convex expected functions 
. 
While convergence is an important property of the any-time methods, arbitrary 
stationary points do not yield approximation guarantees for general non-convex functions.

\citet{Johnson2013Accelerating} introduced the variance reduction 
technique for accelerating stochastic gradient descent. It was 
independently discovered by~\citet{mahdavi2013mixed}.
\citet{allen2016variance} applied this technique to non-convex 
optimization. \citet{hazan2016variance} devised a projection-free 
stochastic convex optimization algorithm based on this technique.
\citet{mokhtari2017conditional,mokhtari2018stochastic} proposed the first sample-efficient variance reduction technique for projection-free algorithms that does not require increasing batch sizes. Their method achieves the tight $ (1-1/e)$ approximation guarantee for monotone and continuous DR-submodular functions. 
Although these variance reduction techniques have enjoyed success 
in the offline setting, they have yet to be as extensively applied in the online setting that we consider in 
this paper.

In the discrete domain, \citet{streeter2009online} studied the online maximization problem of 
monotone submodular set functions subject to a knapsack constraint 
and introduced the meta-action technique.
In a celebrated work, \citet{calinescu2011maximizing} proposed an (offline) method for maximizing 
monotone submodular set functions subject to a matroid constraint by working in the continuous 
domain via the multilinear extension, then rounding the fractional solution. By combining the 
meta-action and lifting techniques, \citet{golovin14online} presented an 
algorithm whose $ (1-1/e) 
$-regret is bounded by $O(\sqrt{T})$. The lifting method therein relies on an expensive sampling 
procedure that does not scale favorably to large applications.

\citet{bach2015submodular} demonstrated connections between continuous submodular functions 
and convex functions in the context of minimization. Building upon the continuous greedy algorithm of 
\cite{calinescu2011maximizing}, \citet{bian16guaranteed} proposed an algorithm that achieves a 
$(1-1/e)$-approximation guarantee for maximizing monotone continuous DR-submodular functions 
subject to 
down-closed convex constraints.  
Projected gradient methods were investigated in~\cite{hassani2017gradient} and 
were shown to attain a $ 1/2 $-approximation ratio for
monotone continuous DR-submodular functions.  Very 
recently, \citet{Chen2018Online} borrowed the 
idea of meta-action~\cite{streeter2009online} and proposed several online algorithms for maximizing
monotone continuous DR-submodular functions. However, each of these methods either requires an 
expensive projection step at each iteration or cannot handle stochastic gradient estimates.

\section{Preliminaries}
\label{sec:prelim}
In this work, we are interested in optimizing two classes of functions, namely convex and continuous 
DR-submodular. 
To begin defining continuous submodular functions, we first 
recall the definition of a submodular set function. A real-valued set function $f: 
2^ \ground 
\rightarrow \reals_+$ is \textit{submodular} if 
$$f(A) + f(B) \geq f(A \cup B) + f(A \cap B)$$ 
for all $A, B \subset \ground$.
The notion of submodularity has been extended to continuous domains \cite{wolsey1982analysis, 
vondrak2007submodularity, bach2015submodular}. Consider a function 
$f:\mathcal{X} \rightarrow \reals_+$ where the domain is of the form $\mathcal{X} = \prod_{i=1}^n 
\mathcal{X}_i$ and each 
$\mathcal{X}_i$ is a compact subset of $\reals_+$. We say that $f$ is \textit{continuous submodular} if 
$f$ is continuous and for all $\bx, \by \in \mathcal{X}$, we have 
$$f(\bx) + f(\by) \geq f(\bx \vee \by) + f(\bx \wedge \by)$$ 
where $\bx \vee \by$ and $\bx \wedge \by$ are component-wise maximum and minimum, 
respectively. Note that we have defined both discrete and continuous functions to be nonnegative on 
their respective domains. For efficient maximization, we also require that these functions satisfy a 
diminishing 
returns condition \cite{bian16guaranteed}. We say that $f$ is \emph{continuous DR-submodular} if $f$ 
is 
differentiable and 
$$\nabla f(\bx) \geq \nabla f(\by)$$ 
for all $\bx \leq \by$. The main attraction of 
continuous DR-submodular functions is that they are concave in positive directions; that is, for all $\bx 
\leq \by$, 
$$f(\by) \leq f(\bx) + \langle \nabla f(\bx), \by - \bx \rangle$$ \cite{calinescu2011maximizing, 
bian16guaranteed}. A function 
$f$ is 
\emph{monotone} if $f(\bx) \leq f(\by)$ for all $\bx \leq \by$. A function $f$ is 
\emph{$L$-smooth} if $\| \nabla f(\bx) - \nabla f(\by) \| \leq L \|\bx - \by \|$ for all 
$\bx, \by$.

We now provide a brief introduction to online optimization, referring the interested reader to the 
excellent 
survey of
\cite{hazan2016introduction}. In the online setting, a player seeks to iteratively optimize a sequence of 
functions
$f_1, \dots f_T$ over $T$ rounds. In each round, a player must first 
choose a point 
$\bx_t$ from the constraint set $\constraint$. After playing $\bx_t$, the value of $f_t(\bx_t)$ is 
revealed to the player, along 
with access to the gradient $\nabla f$. Although the 
player does not know the function $f_t$ while choosing $\bx_t$, they may use information of 
previously seen functions to guide their choice. The situation where an arbitrary sequence of functions 
$f_1, 
\dots, f_T$ is presented is known as the \emph{adversarial online setting}. In the adversarial setting, 
the goal of the player is 
to minimize \textit{adversarial regret}, which is defined as
\begin{dmath*}
	\regret_T \triangleq \sum_{t=1}^T f_t(\bx_t) - \inf_{\bx \in \constraint} \sum_{t=1}^T f_t(\bx)
\end{dmath*}
for minimization problems and analogously defined for maximization problems. Intuitively, a player's 
regret is low if the accumulated value of their actions over the $T$ rounds is close to that of the single 
best action in hindsight. Indeed, this is a natural framework for data-intensive applications where the 
entire data may not fit onto a single disk and thus needs to be processed in  $T$ batches. The 
algorithm designer would like to devise a scheme to process the $T$ batches separately in a way that 
is competitive with the best single disk solution. 

A slightly different formulation known as \emph{stochastic online setting} is when the functions are 
chosen $i.i.d.$ from some unknown distribution $f_t \sim \mathcal{D}$. In this case, the player seeks to 
minimize \emph{stochastic regret}, which is defined as
\begin{dmath*}
	\sregret_T \triangleq \sum_{t=1}^T f(\bx_t) - T \cdot \inf_{\bx \in 
	\constraint} f(\bx)
\end{dmath*}
where $f(\bx) = \expect_{f_t \sim \mathcal{D}} [f_t(\bx)]$ denotes the expected 
function. This is a natural 
framework for many statistical and machine learning applications, such as empirical risk minimization, 
where the true objective is unknown but pairs of data points and labels are sampled. While the 
stochastic setting appears ``easier'' than the adversarial 
setting (in the sense that any strategy for the adversarial settings applies to stochastic settings 
and obtains a potentially lower regret), the strategies designed for the stochastic setting may 
be much simpler and more computationally efficient. For both adversarial and stochastic settings, a 
strategy that achieves a regret that is sublinear in $T$ is considered good and 
$O(\sqrt{T})$ regret bounds are optimal for convex functions in both settings. Although convex 
programs can be efficiently 
solved to high accuracy, general non-convex programs cannot be efficiently exactly optimized, thus 
necessitating another definition of regret. The $\alpha$-regret is defined as
\begin{dmath*}
\alpha \mbox{-} \regret_T \triangleq \alpha \sup_{\bx \in \constraint}	\sum_{t=1}^T 
f_t(\bx) - \sum_{t=1}^T f_t(\bx_t) 
\end{dmath*}
for adversarial maximization problems, and may be analogously extended to other scenarios. Intuitively, 
$\alpha$-regret compares a player's actions with the best $\alpha$-approximation to the optimal 
solution in hindsight. This is appropriate when the objective functions do not admit efficient 
optimization routines, but do admit constant-factor approximations, as is the case with continuous 
DR-submodular functions.

Nearly all optimization methods for both offline and online settings use first order information of the 
objective function; however, exact 
gradient computations can be costly, especially when the objective function is only readily expressed 
as a large sum of individual functions or is itself an expectation over an unknown distribution.  In this 
case, stochastic estimates are usually much more 
computationally efficient to obtain via sampling or simulation. In this work, we assume that once a 
function $f_t$ is revealed, the
player gains oracle access to unbiased stochastic estimates of the gradient, rather than the exact 
gradient. More precisely, 
the player may query the oracle to obtain a random linear function $\tnabla f(x)$ such that $\expect 
[ \nabla f(x) - \tnabla f(x) ] = 0$ for all $x$. This computational model captures commonly used 
mini-batch methods for 
estimating gradients, among other examples. In this work, we make a few main assumptions that allow 
our algorithms to be analyzed.

\begin{assumption} \label{assumption:constraint}
	The constraint set $\constraint$ is convex and compact, with diameter $D = \sup_{\bx,\by \in 
	\constraint} \|\bx - \by\|$ and radius $R = \sup_{\bx \in \constraint} \|\bx \|$.
\end{assumption}

\begin{assumption} \label{assumption:smooth}
	In the adversarial setting, each function $f_t$ is $L$-smooth and in the stochastic setting, the 
	expected function $f$ is $L$-smooth.
\end{assumption}

\begin{assumption} \label{assumption:variance}
	In the adversarial setting, the gradient oracle is unbiased
	$\expect[ \nabla f_t(\bx) - \tnabla f_t(\bx) ] = 0$
	and has a bounded variance
	$\expect[ \| \nabla f_t(\bx) - \tnabla f_t(\bx) \|^2 ] \leq \sigma^2$ 
	for all points $\bx$ and functions $f_t$.
	In the stochastic setting, the gradient oracle is unbiased
	$\expect[ \nabla f(\bx) - \tnabla f_t(\bx) ] = 0$
	and has a bounded variance
	$\expect[ \| \nabla f(\bx) - \tnabla f_t(\bx) \|^2 ] \leq \sigma^2$ for all 
	points $\bx$ and functions 
	$f_t$.
\end{assumption}
We remark that in the stochastic setting and under mild regularity conditions, unbiasedness of the 
gradients $\expect[ \nabla f_t(\bx) - \tnabla f_t(\bx) ] = 0$ implies 
unbiasedness $\expect[ \nabla f(\bx) 
- \tnabla f_t(\bx) ] = 0$ in 
Assumption~\ref{assumption:variance} because $f(\bx) = \expect_{f_t \sim 
\mathcal{D}}[ f_t(\bx)]$ 
Moreover, upper bounds on the variance terms $ \expect[ \| \nabla f(\bx) - \nabla 
f_t(\bx) \|^2 ] \leq 
\sigma_a^2 $ and 
$ \expect[ \| \nabla f_t(\bx) - \tnabla f_t(\bx)\|^2 ] \leq \sigma_b^2 $ yield a variance bound of 
$\expect[ \| \nabla f(\bx) - \tnabla f_t(\bx) \|^2 ] \leq \sigma_a^2 + 
\sigma_b^2$
, by the triangle inequality.

\section{Main Results}\label{sec:main_results}

We now present two algorithms for online optimization of convex and continuous DR-submodular 
functions in the 
adversarial and stochastic settings.
Unlike previous work, these methods are projection-free and require only \emph{stochastic 
estimates} of the gradients, rather than exact gradient computations. In both algorithms, the main 
computational primitive is linear optimization over a compact convex set. 
In addition, we remark that both algorithms can be converted into an 
anytime 
algorithm that does not require the knowledge of the horizon $ T $ via the 
doubling trick; see Section~2.3.1 of \cite{shalev2012online}.

\subsection{Adversarial Online Setting}
Algorithm~\ref{alg:meta_frank_wolfe} combines the recent variance reduction technique of 
\cite{mokhtari2017conditional} along with the use of online linear optimization oracles to minimize the 
regret in each round. 
An online linear optimization oracle is an instance of an online linear 
optimization 
(minimization/maximization in the convex/DR-submodular setting, respectively) 
algorithm that optimizes linear objectives in a sequential manner.
Both the variance reduction in the stochastic gradient estimates and the online 
linear oracles 
are crucial in the algorithm, as just one technique is not enough to get sublinear regret bounds in the 
adversarial setting. At a high level, our algorithm produces iterates $x_t$ by running $K$ steps of a 
Frank-Wolfe procedure, using an average of previous gradient estimates and 
linear online optimization 
oracles in 
place of exact optimization of the true gradient. 
After a point $x_t$ is played in round $t$, our 
algorithm 
queries the gradient 
oracle $\tnabla f_t$ at $K$ points. Then, the gradient estimates are averaged with those from previous 
rounds and fed as objective functions into $K$ linear online optimization oracles. The $K$ points 
chosen by the oracles are used as iterates in a full $K$-step Frank-Wolfe 
subroutine to obtain the next 
point $x_{t+1}$. A formal description is provided in Algorithm~\ref{alg:meta_frank_wolfe}.

\begin{algorithm}[htb]
	\begin{algorithmic}[1] 
		\REQUIRE 
		convex set $\constraint$, time 
		horizon $T$, linear optimization oracles $\mathcal{E}^{(1)} \dots \mathcal{E}^{(K)}  $, 
		step sizes $\rho_k \in (0,1)$ and $\eta_k \in (0,1)$, and 
		initial point $\bx_1$
		\ENSURE $\{\mathbf{x}_t:1\leq t\leq T \}$
		\STATE Initialize online linear optimization oracles $\mathcal{E}^{(1)} 
		\ldots \mathcal{E}^{(K)}$
		\STATE Initialize $\bd_t^{(0)} = 0$ and $\bx_t^{(1)} = \bx_1$ 
		\FOR{$t\gets 1,2,3,\ldots, T$} 
		\STATE $\bv_t^{(k)} \gets$ output of oracle $\mathcal{E}^{(k)}$ in round $t-1$
		\STATE $\bx_t^{(k+1)} \gets update(\bx_t^{(k)}, \bv_t^{(k)}, \eta_k)$ for $k=1\dots K$
		\STATE Play $\bx_t = \bx_t^{(K+1)}$, then obtain value $f_t(\bx_t)$ and unbiased 
		oracle access to $\nabla f_t$
		\STATE $\bd_t^{(k)} \gets (1 - \rho_k) \bd_t^{(k-1)} + \rho_k \tnabla 
		f_t(\bx_t^{(k)})$ for $ k=1\ldots K $
		\STATE Feedback $\langle \bv_t^{(k)}, \bd_t^{(k)} \rangle$ to 
		$\mathcal{E}^{(k)}$ for $k=1\dots K$
		\ENDFOR
	\end{algorithmic}\caption{\AlgMetaFW
		\label{alg:meta_frank_wolfe}}
\end{algorithm}

There are only a few differences in Algorithm~\ref{alg:meta_frank_wolfe} for convex and 
submodular optimization. First, the online oracles should be minimizing in the case of convex 
optimization and maximizing in the case of submodular optimization. Second, the initial point $\bx_1$ 
may be any point in $\constraint$ for convex problems but should be set to $0$ for submodular 
problems (even if $\constraint$ is not down-closed). Finally, the update rule is 
$$\bx_t^{(k+1)} \gets (1 - \eta_k) \bx_t^{(k)} + \eta_k \bv_t^{(k)}$$
for convex problems and 
$$\bx_t^{(k+1)} \gets \bx_t^{(k)} + \eta_k \bv_t^{(k)}$$ 
for submodular problems. We now provide a formal regret bound.

\begin{theorem}[Proof 
in~\cref{app:adversarial_submodular,app:adversarial_convex}] 
\label{thm:adversarial}
	Suppose Assumptions  \ref{assumption:constraint} - \ref{assumption:variance} hold, the online 
	linear optimization oracles have regret at most $\oregret$, 
	and the 
	averaging parameters are chosen as $\rho_k = \frac{2}{(k+3)^{2/3}}$.
	Then for convex functions $f_1, \dots, f_T$ and step sizes $\eta_k = 
	\frac{1}{k+3}$, the adversarial 
	regret of Algorithm~\ref{alg:meta_frank_wolfe} is at most
	\begin{dmath*}
		\frac{4 T D Q^{1/2}}{K^{1/3}} + \frac{4 T}{K} \left(M + \frac{LD^2}{3} \log (K +1)
		\right) + \frac{4}{3} \oregret
	\end{dmath*} 
	in expectation, where 
	$M= \max_{1\leq t \leq T} \left[ f_t(\bx_1) - f_t(\bx^*) \right]$ and 
	$Q \triangleq \max \{ 4^{2/3} \max_{1 \leq t \leq T} \|\nabla f_t(\bx_1) \|^2 , 
	4\sigma^2 + 3(LD)^2/2 \}$. For monotone continuous 
	DR-submodular functions $f_1, 
	\dots, f_T$ and 
	step sizes $\eta_k = \frac{1}{K}$, the adversarial $(1-1/e)$-regret of 
	Algorithm~\ref{alg:meta_frank_wolfe} is at most
	\begin{dmath*}
		\frac{3TDQ^{1/2}}{2K^{1/3}}  + \frac{LD^2T}{2K} + \oregret
	\end{dmath*}
	in expectation, where $ Q \triangleq \max\{ \max_{1\leq t\leq 
		T} \lVert \nabla f_t(\bx_1) 
	\rVert^2 4^{2/3}, 4\sigma^2 + 6L^2 R^2 \} $.
\end{theorem}
From \cref{thm:adversarial}, we observe that by setting $K=T^{3/2}$ and 
choosing a projection-free online linear optimization oracle with $\oregret = O(\sqrt{T})$, such as 
Follow the Perturbed Leader \cite{Cohen2015following}, both regrets are bounded above by 
$O(\sqrt{T})$. We 
remark that the expectation in 
Theorem~\ref{thm:adversarial} is with respect to the stochastic gradient estimates.

\subsection{Stochastic Online Setting}

In the stochastic online setting, where functions are sampled i.i.d. $f_t \sim \mathcal{D}$, we can 
develop much simpler algorithms that still achieve sublinear regret. Algorithm~\ref{alg:vr_frank_wolfe} 
works without instantiating any online linear optimization oracles and requires only a single 
stochastic estimate of the gradient at each round. Indeed, because the functions are not arbitrarily 
chosen, variance reduction along with one Frank-Wolfe step suffices to achieve a sublinear regret 
bound.

\begin{algorithm}[htb]
	\begin{algorithmic}[1] 
		\REQUIRE 
        convex set $\constraint$, time 
        horizon $T$, step sizes $\rho_t\in (0,1)$ and $\eta_t \in (0,1)$, and 
        initial point $\bx_1$
		\ENSURE $\{\mathbf{x}_t:1\leq t\leq T \}$
		\STATE $\bd_0\gets 0$
		\FOR{$t\gets 1,2,3, \ldots, T$} 
        \STATE Play $\bx_t$, then obtain value $f_t(\bx_t)$ and unbiased 
        oracle access to $\nabla f_t$
		\STATE $\bd_t \gets (1-\rho_t)\bd_{t-1} + \rho_t \tnabla 
		f_t(\bx_t)$
        \STATE $\bv_t \gets \argmax_{\bv \in \constraint} \langle 
        \bd_t, \bv \rangle$
        \STATE $\bx_{t+1}\gets  update(\bx_t, \bv_t, \eta_t)$
		\ENDFOR
	\end{algorithmic}
	\caption{\AlgOneShotFW 
	\label{alg:vr_frank_wolfe}}
\end{algorithm}

The differences in Algorithm~\ref{alg:vr_frank_wolfe} for convex and submodular optimization are 
similar to those in Algorithm~\ref{alg:meta_frank_wolfe}. Namely, the update rules are the same and the 
initial point $\bx_1$ may be arbitrarily chosen from $\constraint$ for convex optimization, and set to 
$0$ for submodular optimization.

\begin{theorem}[Proof 
in~\cref{app:stochastic_submodular,app:stochastic_convex}] 
\label{thm:stochastic}
	Suppose Assumptions  \ref{assumption:constraint} - \ref{assumption:variance} hold and the 
	averaging parameters are chosen as $\rho_t = \frac{2}{(t+3)^{2/3}}$.
	Then for a convex expected function $f$ and step sizes $\eta_t = 
	\frac{1}{t+3}$, the stochastic 
	regret of Algorithm~\ref{alg:meta_frank_wolfe} is at most
	\begin{dmath*}
		4 M \log(T+1) + 6 Q^{1/2}D T^{2/3} + \frac{4}{3} LD^2 \log^2(T+3)
	\end{dmath*} 
	in expectation, where $M= f(\bx_1) - f(\bx^*)$ and $Q \triangleq \max \{ 
	4^{2/3} \|\nabla F(\bx_1) \|^2 
	, 
	4\sigma^2 + 3(LD)^2/2 \}$. For expected functions $f$ 
	which are monotone 
	continuous 
	DR-submodular and 
	step sizes $\eta_k = \frac{1}{K}$, the stochastic $(1-1/e)$-regret of 
	Algorithm~\ref{alg:vr_frank_wolfe} is at most
	\begin{dmath*}
		(1-1/e)M + \frac{3DQ^{1/2}}{10}(3T^{2/3}+2T^{-1})+\frac{LD^2}{2}.
	\end{dmath*}
	in expectation, where $M=f(\bx^*)-f(0)$ and $Q \triangleq \max\{ 
	\lVert \nabla f(0) 
	\rVert^2 4^{2/3}, 4\sigma^2 + 6L^2 R^2 \}$
\end{theorem}

\subsection{Lifting Methods for Discrete Online Optimization}

One exciting application of our online continuous DR-submodular optimization algorithms is a 
new approach for online discrete submodular optimization. While previous methods could only handle 
knapsack constraints \cite{streeter2009online} or required expensive sampling procedures 
\cite{golovin14online}, our continuous methods can be applied to the discrete setting to handle general 
matroid constraints and computationally cheap sampling procedures.

Suppose $f_1, \dots f_T$ are nonnegative monotone submodular set functions on a ground set 
$\Omega$ with matroid constraint $\setsys$ and $\bar{f}_1 , 
\dots \bar{f}_T$ are corresponding multi-linear extensions with matroid 
polytope $\constraint \subset 
[0,1]^n$. A 
discrete procedure that uses our continuous algorithm is as follows: at 
each round $t$, the online continuous algorithm produces a fractional solution 
$\bx_t \in \constraint 
$, which 
is then 
rounded to a set $X_t \in \setsys$ and played as the discrete solution. The value $f_t(X_t)$ is 
revealed and the  
player is granted access to the discrete function $f_t$. Then, the player supplies the 
continuous algorithm with a stochastic gradient estimate $\tnabla \hat{f}_t$ obtained \emph{by a 
single function evaluation}, as 
\begin{equation}\label{eq:gradient}
	\frac{\partial f_t(\bx) }{\partial x_i} = \expect[f(R\cup\{i\} )-f(R)], 
	\enspace 
	\forall i\in [n],
\end{equation}
where $ R $ is random subset of $ [n]\setminus \{i\} $ such that for every $ 
j\in [n]\setminus \{i\} $, the event $ j\in R $ happens with an independent 
probability 
of $ x_j $.
 Because a lossless rounding scheme is used, the discrete player enjoys a 
 regret that is no worse than 
that of the continuous solution. Provably lossless rounding schemes include the pipage 
rounding~\cite{ageev2004pipage,calinescu2011maximizing} and contention 
resolution~\cite{vondrak2011submodular}. 

\begin{figure}[htb]
	\centering
	\begin{framed}
		\begin{tikzpicture}
		\matrix (m) [matrix of  nodes,row sep=3em,column 
		sep=3.5em,minimum 
		width=2em]
		{
			Multilinear extension  & Continuous 
				solution \\
			Submodular set fn. & Discrete 
			solution 
			\\};
		\path[-stealth]
		(m-2-1) edge node [right] {One sample} (m-1-1)
		(m-1-1) edge node [above] {Stochastic optimization 
		algorithm 
			$\mathcal{A}$} 
		(m-1-2)
		(m-1-2) edge node [left] {Rounding} (m-2-2);
		\end{tikzpicture}
	\end{framed}
	\caption{Diagram of the one-sample lifting 
	method\label{fig:diagram} }
\end{figure}
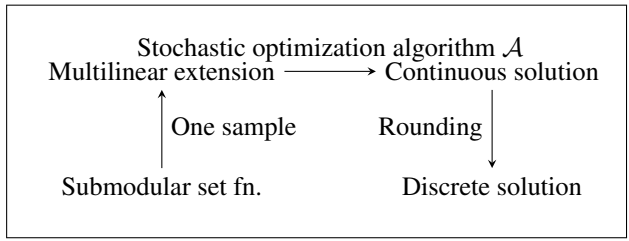

Most discrete submodular maximization algorithms that go through the multi-linear extension require a 
gradient estimate with high accuracy. In order to do this, they appeal to a concentration bound, which 
requires $O(n^2)$ evaluations of the discrete function for independently chosen samples. In stark 
contrast, our algorithms can handle stochastic gradient estimates and thus require 
only a single function evaluation, finally making continuous methods a reality for large-scale online 
discrete optimization problems. The framework of 
the one-sampling lifting method is illustrated in~\cref{fig:diagram}.

As an example, we present in~\cref{alg:lifting} how to use \AlgMetaFW as an 
online maximization 
algorithm of submodular set functions. According to \cref{thm:adversarial}, the 
$ (1-1/e) $-regret of \cref{alg:lifting} is bounded by 
$	\frac{3TDQ^{1/2}}{K^{1/3}}  + \frac{LD^2T}{2K} + \oregret$,
where $ \oregret $ is the regret of $ \mathcal{E}^{(k)} $ up to horizon $ T $. 
If one sets $ \mathcal{E}^{(k)} $ to an online linear maximization algorithm 
with regret bound $ O(\sqrt{T}) $ and sets $ K=T^{3/2} $, the $ (1-1/e) 
$-regret is at most $ O(\sqrt{T}) $.
\begin{algorithm}[htb]
	\begin{algorithmic}[1] 
		\REQUIRE 
		matroid constraint $ \mathcal{I} $, time 
		horizon $T$, linear optimization oracles $\mathcal{E}^{(1)} \dots \mathcal{E}^{(K)} $, 
		step sizes $\rho_k \in (0,1)$ and $\eta_k \in (0,1)$, and 
		initial point $\bx_1$
		\ENSURE $\{X_t:1\leq t\leq T \}$
		\STATE Initialize online linear optimization oracles $\mathcal{E}^{(1)} 
		... \mathcal{E}^{(K)}$, setting the constraint set to the matroid 
		polytope of $ \mathcal{I} $
		\STATE Initialize $\bd_t^{(0)} = 0$ and $\bx_t^{(1)} = \bx_1$ 
		\FOR{$t\gets 1,2,3,\ldots, T$} 
		\STATE $\bv_t^{(k)} \gets$ output of oracle $\mathcal{E}^{(k)}$ in 
		round $t-1$
		\STATE $\bx_t^{(k+1)} \gets update(\bx_t^{(k)}, \bv_t^{(k)}, \eta_k)$ 
		for $k=1\ldots K$
		\STATE  $\bx_t \gets \bx_t^{(K+1)}$
		\STATE play $ X_t\gets round(\bx_t) $, obtain value $f_t(X_t)$ and 
		observe
		the function $ f_t $
		\STATE Sample $ \tnabla \bar{f}_t(\bx_t^{(k)}) $ for $ k=0,\ldots,K-1 
		$
		\STATE $\bd_t^{(k)} \gets (1 - \rho_k) \bd_t^{(k-1)} + \rho_k \tnabla 
		f_t(\bx_t^{(k)})$ for $ k=1\ldots K $
		\STATE Feedback $\langle \bv_t^{(k)}, \bd_t^{(k)} \rangle$ to 
		$\mathcal{E}^{(k)}$ for $k=1\ldots K$
		\ENDFOR
	\end{algorithmic}\caption{\AlgMetaFW for online discrete submodular 
	maximization
		\label{alg:lifting}}
\end{algorithm}

\section{Experiment}\label{sec:experiment}
In this section, we test our online algorithms for monotone continuous DR-submodular and convex 
optimization on both real-world and synthetic data sets. We find that our algorithms outperform most 
baselines, including projected gradient descent, when supplied with stochastic gradient estimates. All 
code was 
written in the Julia programming language and tested on a Macintosh desktop with an Intel 
Processor i7 with 16 GB of RAM. No parts of the code were optimized past basic 
Julia usage.
A list of all algorithms to be compared in this section is presented below.
\begin{itemize}
	\item Meta-Frank-Wolfe is \cref{alg:meta_frank_wolfe}. We compare the 
	variance-reduced 
	meta-Frank-Wolfe 
	algorithm and the analogue without variance 
	reduction, denoted \AlgMetaVR and \AlgMetaNVR, respectively.
	\item One-shot Frank-Wolfe is \cref{alg:vr_frank_wolfe}. We compare the 
	One-shot online 
	Frank-Wolfe algorithm with and without variance reduction, denoted 
	\AlgOnlineVR \AlgOnlineNVR, respectively.
	\item Regularized online Frank-Wolfe is referred to as the online 
	conditional gradient algorithm in~\cite{hazan2016introduction}. It has a 
	regularizer term when computing the gradient. Thus we term it the 
	regularized online Frank-Wolfe algorithm and denote it as \AlgRegu.
	\item Online projected gradient ascent (\AlgOGA) follows the 
	direction of the 
	projected gradient. Its $ 1/2 $-regret is 
	at most $ O(\sqrt{T}) $ for online monotone continuous DR-submodular 
	maximization if the step size is set to $ \Theta(1/\sqrt{t}) 
	$ on the $ t $-th iteration~\cite{Chen2018Online}. Note that \AlgOGA is not 
	a projection-free algorithm. In the setting of convex minimization, we use online projected 
	gradient descent instead (denoted by \AlgOGD).
	\item When we perform experiments on discrete submodular maximization 
	problems using our lifting method, we also compare the above algorithms with the 
	\AlgOG algorithm \cite{streeter2009online}. 
\end{itemize}

\begin{figure*}[htb]
%
%
\centering
	\subfigure[Continuous facility location on Jester dataset
	\label{fig:jester}]{\includegraphics[width=0.33\textwidth]{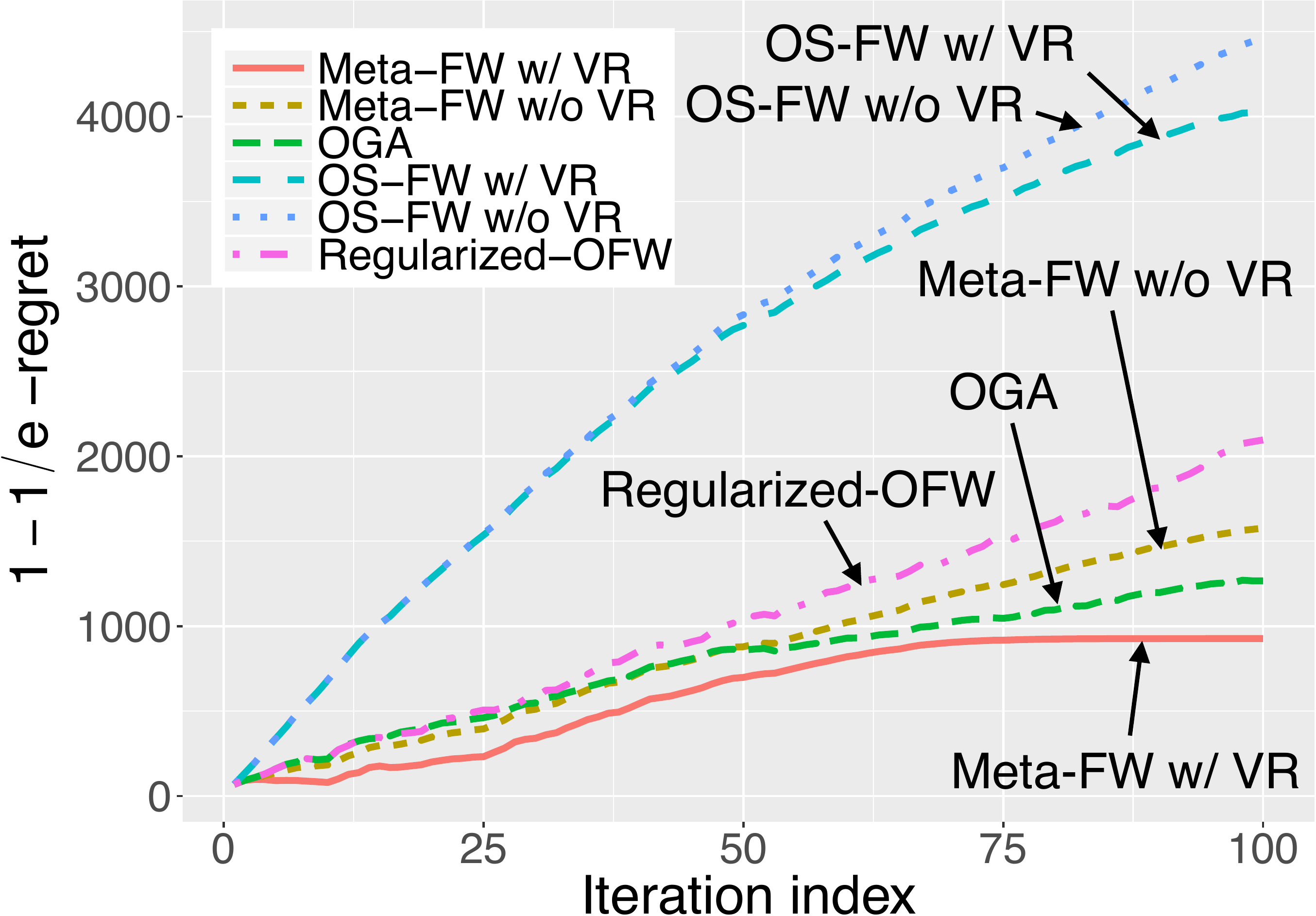}}
	\subfigure[Discrete facility location on Jester dataset
	\label{fig:lifting}]{\includegraphics[width=0.33\textwidth]{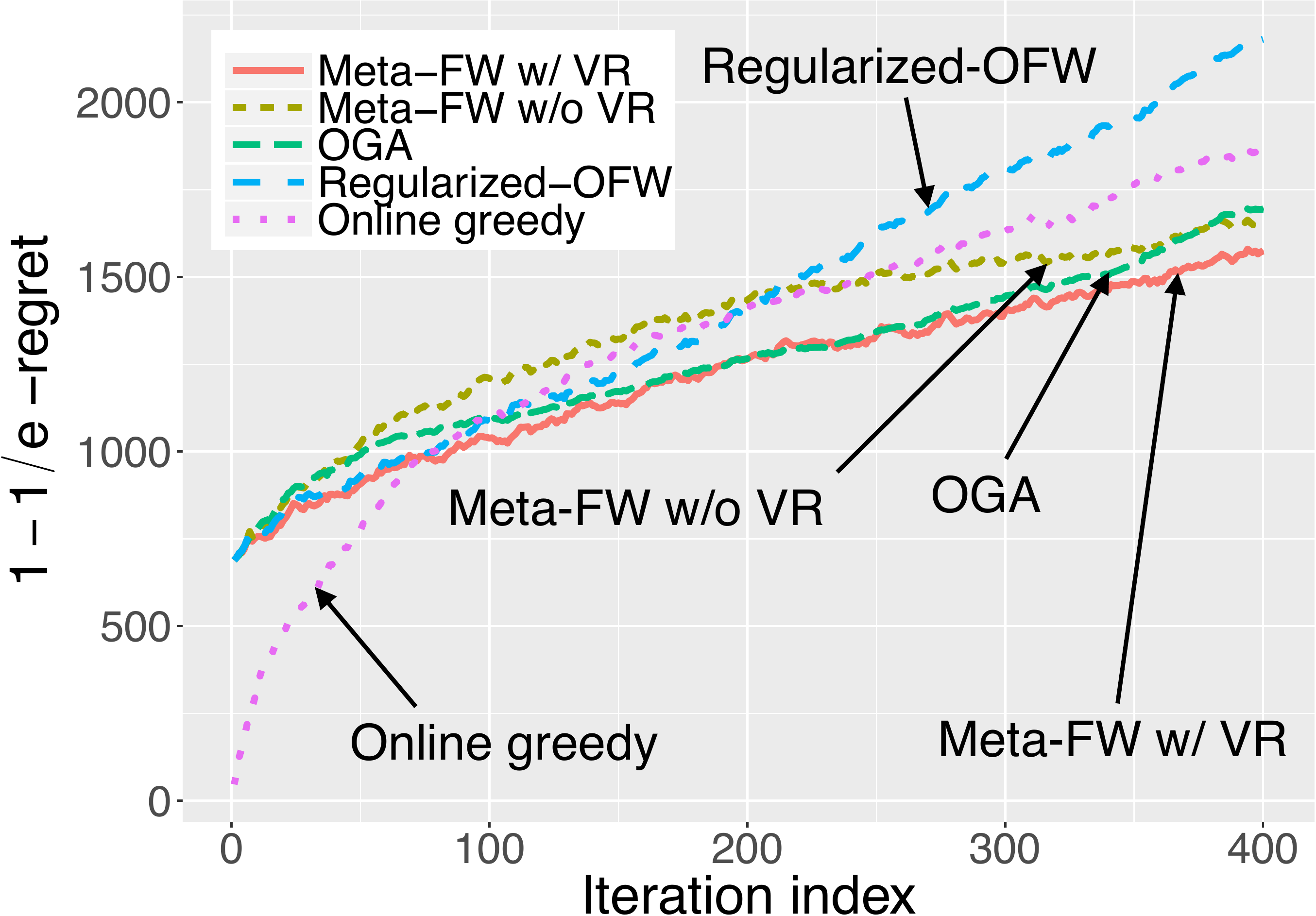}}
	\subfigure[News recommendation in Reuters corpus
	\label{fig:reuters}]{\includegraphics[width=0.33\textwidth]{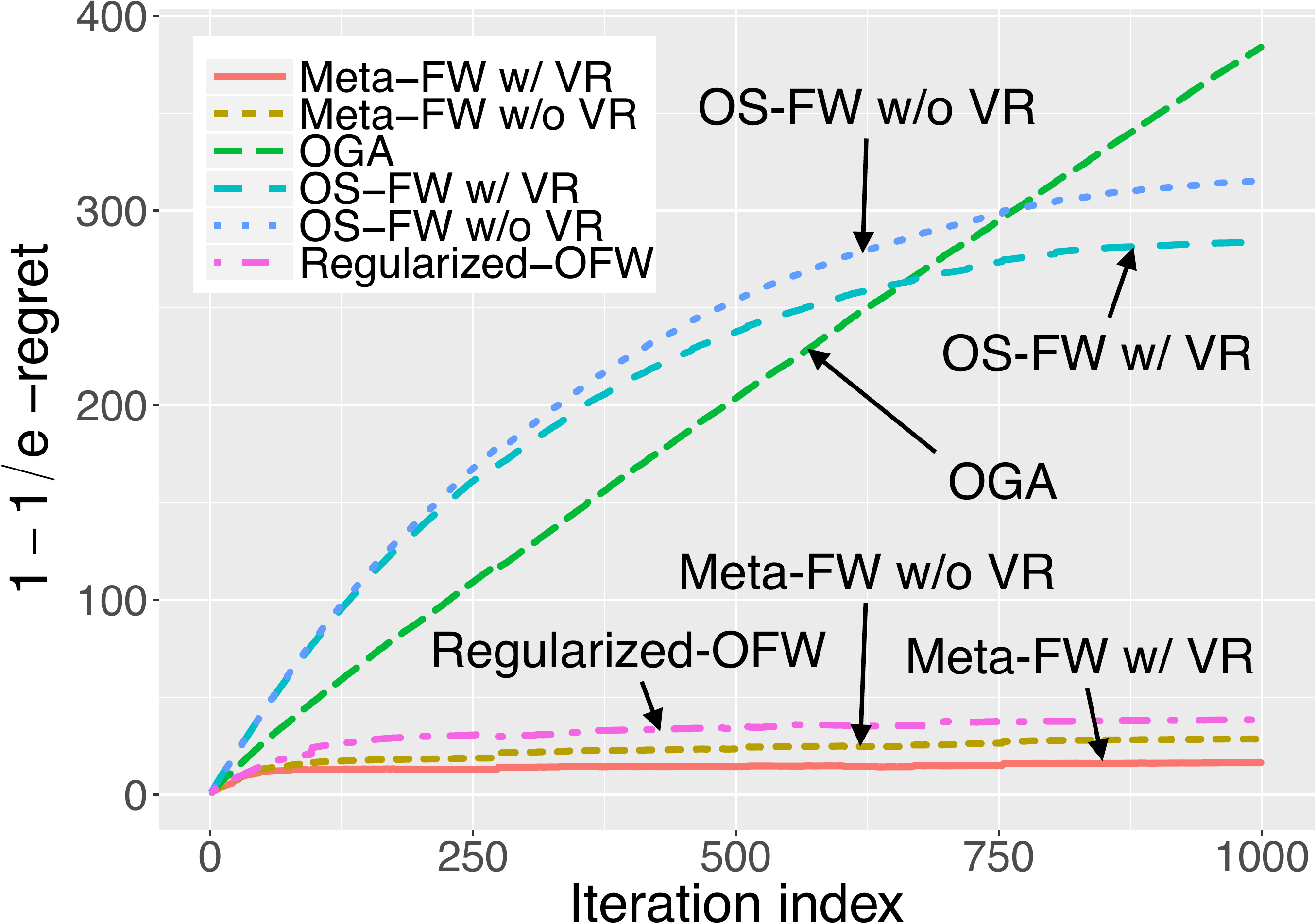}}

	%
	%
	\subfigure[Network flow
	\label{fig:minflow}]{\includegraphics[width=0.33\textwidth]{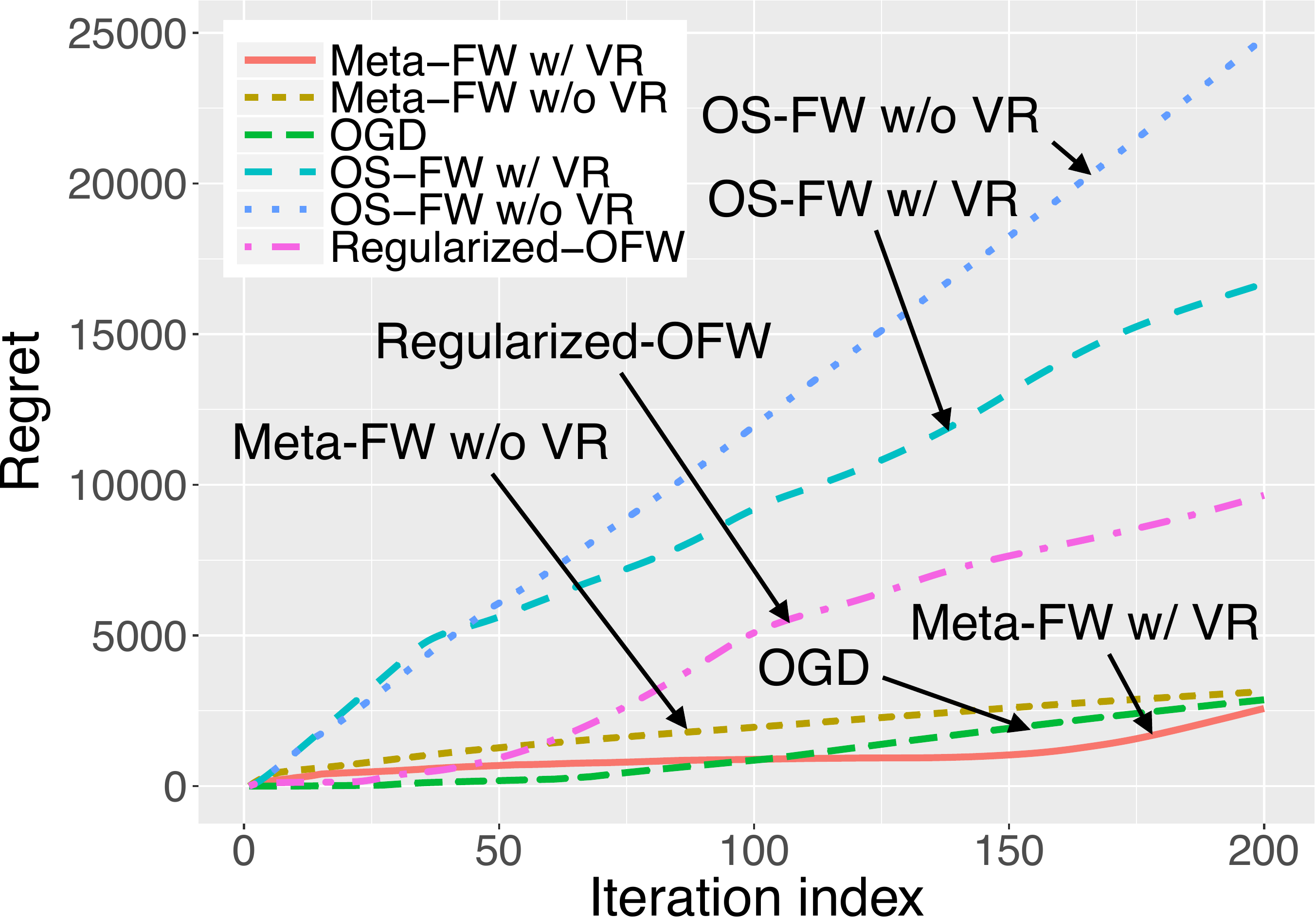}}\subfigure[Matrix
	 completion
	\label{fig:matrix}]{\includegraphics[width=0.33\textwidth]{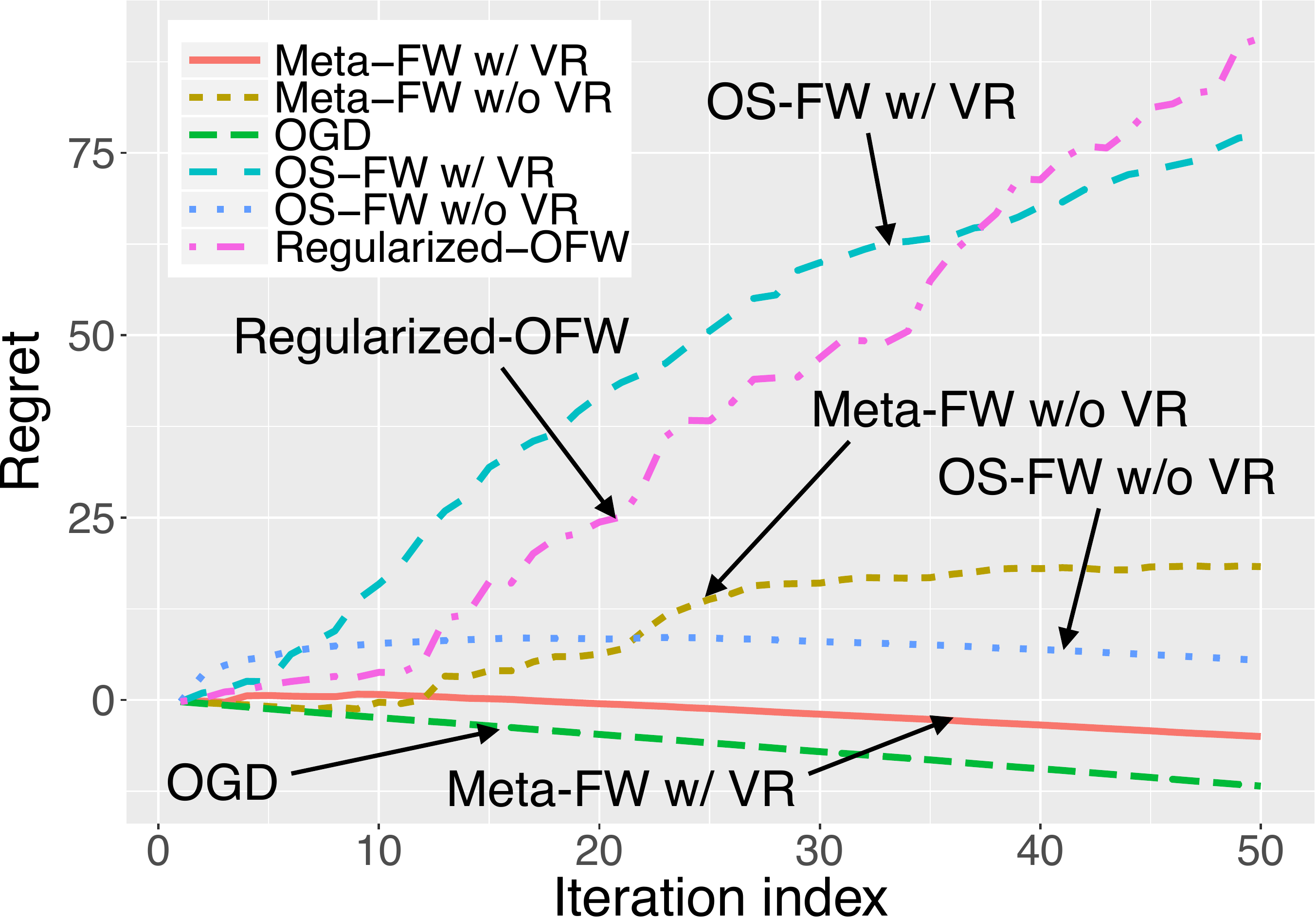}}
	\subfigure[Execution time of matrix completion
	\label{fig:time}]{\includegraphics[width=0.33\textwidth]{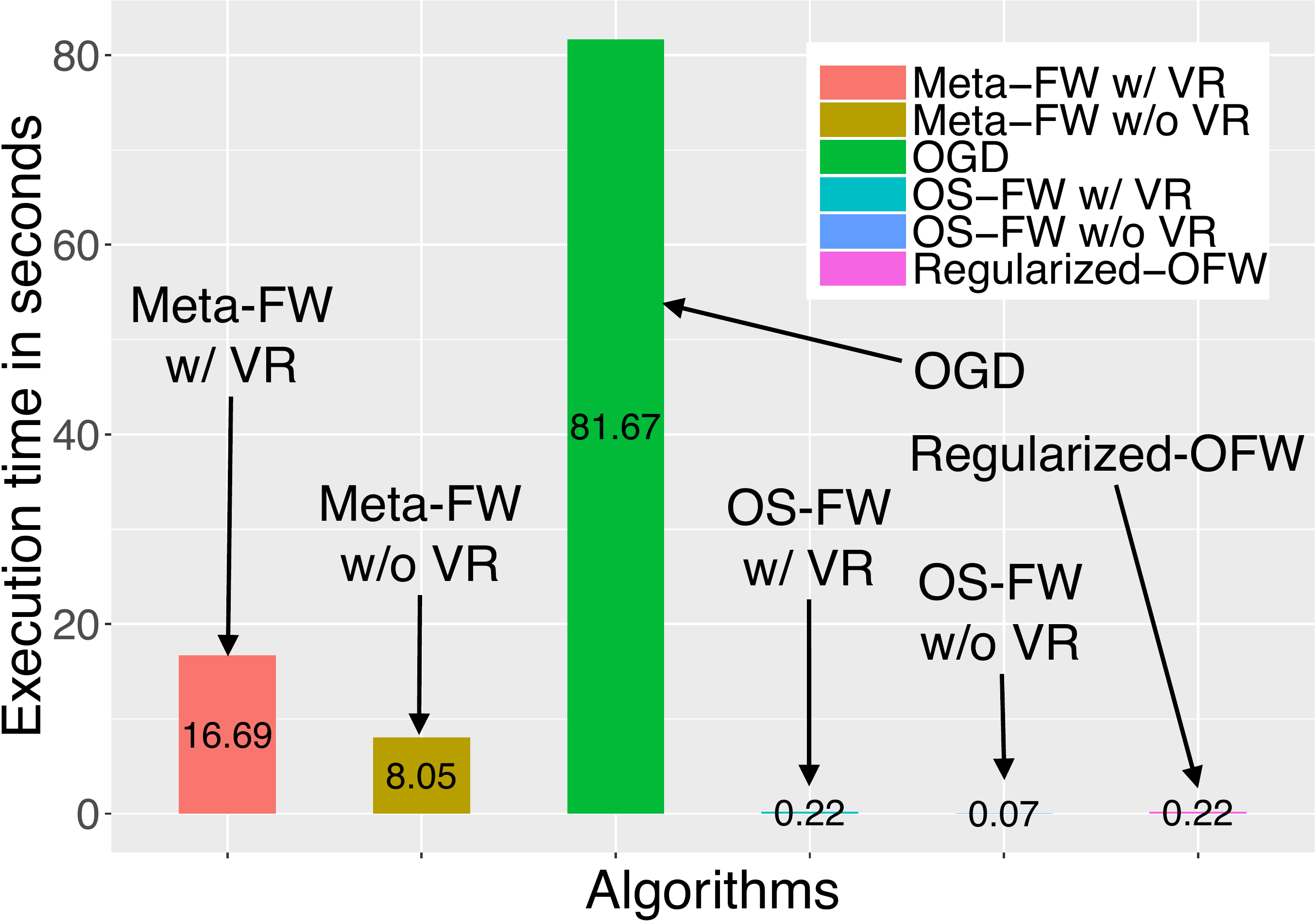}}
\caption{
	\cref{fig:jester,fig:lifting,fig:reuters} shows the $ (1-1/e) $-regret 
of online DR-submodular maximization algorithms. 
 In~\cref{fig:jester,fig:lifting}, we show the 
regret for the continuous and discrete facility location 
objective 
functions on the Jester dataset, respectively. 
 We show the results for the online news recommendation problem in the Reuters 
 corpus
in~\cref{fig:reuters}.
The results for online convex minimization are illustrated 
in~\cref{fig:minflow,fig:matrix,fig:time}.
In~\cref{fig:minflow}, we show the regret of the 
algorithms applied to the stochastic cost network flow problem. The results for 
the matrix completion problem are shown
in~\cref{fig:matrix} and the computational time is 
illustrated in~\cref{fig:time}.
}
\end{figure*}

\subsection{Online DR-Submodular Maximization}
In order to test the performance of algorithms for online maximization of 
monotone continuous DR-submodular functions with stochastic gradient estimates, we conducted 
three 
sets of experiments on real-world datasets. We approximate the $(1-1/e)$-regret by running an offline 
Frank Wolfe maximization to produce a solution that is a $(1-1/e)$ approximation to the 
optimum.

\paragraph{Joke Recommendations (Continuous)} The first set of experiments is to optimize a 
sequence of continuous facility 
location 
objectives on the Jester dataset~\cite{Goldberg2001Eigentaste:}. It contains 
ratings of 100 jokes from 73,421 users and the rating range is $ [-10, 10] $. 
We re-scale the rating range into $ [0, 20] $ so that all ratings are 
non-negative. Let $ R_{uj} $ be user $ u $'s rating of joke $ j $. All users 
are splitted into disjoint batches $ B_1, B_2, \ldots, B_T $, each containing $ 
B $ users. The facility location objective is defined as $ f_t(X) = \sum_{u\in 
B_t} \max_{j\in X} R_{uj},\enspace \forall X \subseteq [J]$,  
where $ J=100 $ is the total number of jokes and $ [J]=\{1,2,3,\ldots, J \} $. 
Its multilinear extension is given by $ 
\bar{f}_t(\bx) = \sum_{u\in B_t} \sum_{l=1}^{J} R_{u j_u^l} \bx_{j_u^l} 
\prod_{m=1}^{l-1} (1-\bx_{j_u^m} ), \enspace \forall \bx\in [0,1]^J$, where 
$ 
j^1_u, j^2_u,\dots, j^J_u $ is a permutation of $ 1,2,\dots, J $ such that $ 
R_{uj^1_u}\ge R_{uj^2_u} \ge \dots \ge R_{uj^J_u} $~\cite{Iyer2014Monotone}.
In this experiment, the sequence of objective functions to be optimized is $\{ \bar{f}_1, 
\bar{f}_2, \ldots, \bar{f}_T \}  $. 
The stochastic gradient is obtained by the sampling method given 
in~\cref{eq:gradient} with only one sample for each coordinate of the gradient.
We set the constraint set to $ \{ \bx\in 
[0,1]^J: \bm{1}^\top \bx \leq 1 \} $ and choose $ B=5 $. We present  the 
results in~\cref{fig:jester}. 
\AlgMetaVR attains the smallest 
regret. The counterpart without variance reduction \AlgMetaNVR is inferior to 
\AlgMetaVR in terms of the regret. 
 \AlgOnlineVR 
outperforms \AlgOnlineNVR, which suggests that the variance reduction technique 
improves the performance of the algorithms.

\paragraph{Joke Recommendations (Discrete)} In the second set experiments, we consider online 
maximization of discrete submodular functions. The problem set up is the same as before, but instead 
of evaluating regret of the multilinear extensions, we round solutions using pipage rounding and 
evaluate the regret on the discrete submodular functions. We set the batch size 
$ B $ to 40 and we recommend 10 jokes for users. The results are illustrated 
in~\cref{fig:lifting}. We observe that \AlgMetaVR outperforms all other 
algorithms again. The projected algorithm \AlgOGA is second to \AlgMetaVR. 
\AlgOG appears only better than \AlgRegu. The experiment result show that the 
continuous algorithms designed under the framework of the lifting method 
perform better than the discrete algorithms.

\paragraph{Topic Summarization} We consider the problem of 
selecting news 
documents in order to maximize the probabilistic coverage of news 
topics~\cite{el2009turning,Yue2011Linear}.
 We applied the latent Dirichlet allocation to the 
corpus of Reuters-21578, Distribution 1.0, set the number of topics to 10, and 
extracted the topic distribution 
of each news document. We sample $ T $ batches of news documents from the 
corpus and denote them by $ B_1, B_2, \ldots, B_T $, where each batch contains 
50 randomly sampled documents. For each batch $ B_i $, we define the 
probabilistic coverage function as follows $ f_i(X) = \frac{1}{10} 
\sum_{j=1}^{10}[1-\prod_{a\in X} (1-p_a(j)) ], \enspace \forall X\subseteq B_i, 
$
where $ p_a(\cdot) $ is the topic distribution of news document $ a $. Its 
multilinear extension is $
\bar{f}_i(\bx) = \frac{1}{10} \sum_{j=1}^{10} [ 1-\prod_{a\in X} (1- 
p_a(j)x_a ) ], \enspace \forall \bx\in [0,1]^{50}$, see \cite{Iyer2014Monotone}. The sequence of 
 objective functions that the algorithms are expected to maximize is $ 
 \bar{f}_1, \bar{f}_2, \ldots, \bar{f}_T $. 
 As in the experiments on joke recommendations, the stochastic gradient is 
 obtained by the sampling method given 
 in~\cref{eq:gradient} with only one sample for each coordinate of the gradient.
 The constraint set is $ \{ \bx\in 
 [0,1]^{50}: \bm{1}^\top \bx \leq 45 \} $.
We show the $ (1-1/e) $-regret of the algorithms in~\cref{fig:reuters}. Again, 
\AlgMetaVR exhibits the lowest regret than any other algorithm. Its 
non-variance-reduced counterpart \AlgMetaNVR is second to it. \AlgOnlineVR 
outperforms \AlgOnlineNVR, which confirms the improvement brought by the 
variance reduction technique. 

\subsection{Online Convex Minimization}
The next two sets of experiments test the performance of the 
algorithms for online minimization of convex functions with stochastic gradient estimates. For these 
experiments,  the regret is computed by obtaining the offline solutions with a Frank-Wolfe solver.

\paragraph{Stochastic Cost Network Flow} The fourth set of experiments is a minimum stochastic 
cost flow in a directed network. A directed 
graph $G=(V,E)$ with source $s \in V$, sink $v \in V$, and edge capacities $c : E \rightarrow \reals_+$ 
is known to the player. A \emph{flow} is a function $\bx: \reals_+^{|E|} \rightarrow \reals_+$ 
that satisfies the capacities on each edge  $0 \leq \bx(e) \leq c(e)$ and obeys the conservation laws for 
all vertices $z$,
\[   
\sum_{ \{z,r\} \in E} \bx(r) = 
\begin{cases}
a & z = s\\
-a & z = v \\
0 & \text{otherwise}\\
\end{cases}
\]
for some fixed $a \geq 0$. In each round $t$, a convex cost function on the flow $f_t: \reals^{|E|} 
\rightarrow 
\reals_+$ is drawn 
from a distribution, unknown to the player. The goal is to minimize the stochastic regret of the flows 
chosen. Linear optimizations for this problem may be implemented as combinatorial network flow 
algorithms. 
We used the directed Zachary Karate network with 34 nodes 
and 78 arcs
\cite{zachary1977information}.
 We
set all edge capacities to $1$ and 
cost functions are of the form $f(\bx) = \sum_{e \in E} w_e \bx(e)^2$ where 
$w_e \sim \mathrm{Unif}[100,120]$. 
  The results are presented
in~\cref{fig:minflow}. \AlgMetaVR attains the 
lowest regret among all baselines. Again, the regret of \AlgMetaNVR is larger 
than the variance-reduced \AlgMetaVR. Similarly, \AlgOnlineVR also outperforms 
\AlgOnlineNVR.

\paragraph{Matrix Completion} 
In the online convex 
matrix completion problem, one would 
like to construct a low rank matrix $X \in \reals^{m \times n}$ that 
well-approximates a given matrix $M 
\in \reals^{m \times n}$ on observed entries 
$OB \subseteq [m] \times [n]$. The convex relaxation is
$
	\min_{ \Tr(X) \leq k} \sum_{(i,j) \in OB} (X_{i,j} - M_{i,j})^2.
$
In the online setting, observed entries of the matrix arrive in $T$ batches, $OB_1, OB_2, \dots OB_T$, 
each of size $B$. In each round, we construct a low-rank matrix to minimize the total regret over the 
$T$ rounds. Although 
projection involves a full singular value decomposition, linear 
optimization here is simply a calculation of the largest singular vectors of $(X - M)_{OB}$, see Chapter 
7 of \cite{hazan2016introduction}. In our experiment, $M$ is a rank 
10 matrix with 
$m=n=50$, and $B=100$. We illustrate the results in~\cref{fig:matrix}
and  the computational time is shown in \cref{fig:time}. 
\AlgMetaVR is only second to \AlgOGD. However, \AlgOGD is the slowest 
algorithm due to the computationally expensive projection operations and its 
computational time is five times that of  
 \AlgMetaVR. The 
non-variance-reduced \AlgMetaNVR is inferior to \AlgMetaVR in terms of regret.

\section*{Acknowledgments}

AK was supported by AFOSR YIP (FA9550-18-1-0160).
CH was supported in part by 
NSF
GRFP
(DGE1122492) and by ONR Award N00014-16-1-2374.

\clearpage
\bibliography{example_paper}

\begin{thebibliography}{34}
\providecommand{\natexlab}[1]{#1}
\providecommand{\url}[1]{\texttt{#1}}
\expandafter\ifx\csname urlstyle\endcsname\relax
  \providecommand{\doi}[1]{doi: #1}\else
  \providecommand{\doi}{doi: \begingroup \urlstyle{rm}\Url}\fi

\bibitem[Agarwal et~al.(2006)Agarwal, Hazan, Kale, and
  Schapire]{Agarwal2006Algorithms}
Agarwal, A., Hazan, E., Kale, S., and Schapire, R.~E.
\newblock Algorithms for portfolio management based on the newton method.
\newblock In \emph{Proceedings of the 23rd International Conference on Machine
  Learning}, ICML '06, pp.\  9--16, 2006.

\bibitem[Ageev \& Sviridenko(2004)Ageev and Sviridenko]{ageev2004pipage}
Ageev, A.~A. and Sviridenko, M.~I.
\newblock Pipage rounding: A new method of constructing algorithms with proven
  performance guarantee.
\newblock \emph{Journal of Combinatorial Optimization}, 8\penalty0
  (3):\penalty0 307--328, 2004.

\bibitem[Allen-Zhu \& Hazan(2016)Allen-Zhu and Hazan]{allen2016variance}
Allen-Zhu, Z. and Hazan, E.
\newblock Variance reduction for faster non-convex optimization.
\newblock In \emph{International Conference on Machine Learning}, pp.\
  699--707, 2016.

\bibitem[Bach(2015)]{bach2015submodular}
Bach, F.
\newblock Submodular functions: from discrete to continous domains.
\newblock \emph{arXiv preprint arXiv:1511.00394}, 2015.

\bibitem[Bian et~al.(2017)Bian, Mirzasoleiman, Buhmann, and
  Krause]{bian16guaranteed}
Bian, A., Mirzasoleiman, B., Buhmann, J.~M., and Krause, A.
\newblock Guaranteed non-convex optimization: Submodular maximization over
  continuous domains.
\newblock In \emph{AISTATS}, February 2017.

\bibitem[Calandriello et~al.(2017)Calandriello, Lazaric, and
  Valko]{Calandriello2017Second}
Calandriello, D., Lazaric, A., and Valko, M.
\newblock Second-order kernel online convex optimization with adaptive
  sketching.
\newblock In \emph{International Conference on Machine Learning}, ICML '17,
  2017.

\bibitem[Calinescu et~al.(2011)Calinescu, Chekuri, P{\'a}l, and
  Vondr{\'a}k]{calinescu2011maximizing}
Calinescu, G., Chekuri, C., P{\'a}l, M., and Vondr{\'a}k, J.
\newblock Maximizing a monotone submodular function subject to a matroid
  constraint.
\newblock \emph{SIAM Journal on Computing}, 40\penalty0 (6):\penalty0
  1740--1766, 2011.

\bibitem[Chen et~al.(2018)Chen, Hassani, and Karbasi]{Chen2018Online}
Chen, L., Hassani, H., and Karbasi, A.
\newblock Online continuous submodular maximization.
\newblock In \emph{AISTATS}, pp.\  to appear, 2018.

\bibitem[Cohen \& Hazan(2015)Cohen and Hazan]{Cohen2015following}
Cohen, A. and Hazan, T.
\newblock Following the perturbed leader for online structured learning.
\newblock In \emph{Proceedings of the 32Nd International Conference on
  International Conference on Machine Learning - Volume 37}, ICML'15, pp.\
  1034--1042, 2015.

\bibitem[El-Arini et~al.(2009)El-Arini, Veda, Shahaf, and
  Guestrin]{el2009turning}
El-Arini, K., Veda, G., Shahaf, D., and Guestrin, C.
\newblock Turning down the noise in the blogosphere.
\newblock In \emph{SIGKDD}, pp.\  289--298. ACM, 2009.

\bibitem[Frank \& Wolfe(1956)Frank and Wolfe]{frank1956algorithm}
Frank, M. and Wolfe, P.
\newblock An algorithm for quadratic programming.
\newblock \emph{Naval Research Logistics (NRL)}, 3\penalty0 (1-2):\penalty0
  95--110, 1956.

\bibitem[Garber \& Hazan(2013)Garber and Hazan]{garber2013linearly}
Garber, D. and Hazan, E.
\newblock A linearly convergent conditional gradient algorithm with
  applications to online and stochastic optimization.
\newblock \emph{arXiv preprint arXiv:1301.4666}, 2013.

\bibitem[Goldberg et~al.(2001)Goldberg, Roeder, Gupta, and
  Perkins]{Goldberg2001Eigentaste:}
Goldberg, K., Roeder, T., Gupta, D., and Perkins, C.
\newblock Eigentaste: A constant time collaborative filtering algorithm.
\newblock \emph{information retrieval}, 4\penalty0 (2):\penalty0 133--151,
  2001.

\bibitem[Golovin et~al.(2014)Golovin, Krause, and Streeter]{golovin14online}
Golovin, D., Krause, A., and Streeter, M.
\newblock Online submodular maximization under a matroid constraint with
  application to learning assignments.
\newblock Technical report, arXiv, 2014.

\bibitem[Hassani et~al.(2017)Hassani, Soltanolkotabi, and
  Karbasi]{hassani2017gradient}
Hassani, H., Soltanolkotabi, M., and Karbasi, A.
\newblock Gradient methods for submodular maximization.
\newblock \emph{arXiv preprint arXiv:1708.03949}, 2017.

\bibitem[Hazan \& Kale(2012)Hazan and Kale]{hazan2012projection}
Hazan, E. and Kale, S.
\newblock Projection-free online learning.
\newblock In \emph{ICML}, pp.\  1843--1850, 2012.

\bibitem[Hazan \& Luo(2016)Hazan and Luo]{hazan2016variance}
Hazan, E. and Luo, H.
\newblock Variance-reduced and projection-free stochastic optimization.
\newblock In \emph{ICML}, pp.\  1263--1271, 2016.

\bibitem[Hazan et~al.(2016)]{hazan2016introduction}
Hazan, E. et~al.
\newblock Introduction to online convex optimization.
\newblock \emph{Foundations and Trends{\textregistered} in Optimization},
  2\penalty0 (3-4):\penalty0 157--325, 2016.

\bibitem[Iyer et~al.(2014)Iyer, Jegelka, and Bilmes]{Iyer2014Monotone}
Iyer, R., Jegelka, S., and Bilmes, J.
\newblock Monotone closure of relaxed constraints in submodular optimization:
  Connections between minimization and maximization.
\newblock In \emph{Uncertainty in Artificial Intelligence (UAI)}, Quebic City,
  Quebec Canada, July 2014. AUAI.

\bibitem[Johnson \& Zhang(2013)Johnson and Zhang]{Johnson2013Accelerating}
Johnson, R. and Zhang, T.
\newblock Accelerating stochastic gradient descent using predictive variance
  reduction.
\newblock In \emph{NIPS}, pp.\  315--323, 2013.

\bibitem[Lafond et~al.(2015)Lafond, Wai, and Moulines]{lafond2015online}
Lafond, J., Wai, H.-T., and Moulines, E.
\newblock On the online {Frank-Wolfe} algorithms for convex and non-convex
  optimizations.
\newblock \emph{arXiv preprint arXiv:1510.01171}, 2015.

\bibitem[Mahdavi et~al.(2013)Mahdavi, Zhang, and Jin]{mahdavi2013mixed}
Mahdavi, M., Zhang, L., and Jin, R.
\newblock Mixed optimization for smooth functions.
\newblock In \emph{NIPS}, pp.\  674--682, 2013.

\bibitem[Mokhtari et~al.(2018{\natexlab{a}})Mokhtari, Hassani, and
  Karbasi]{mokhtari2017conditional}
Mokhtari, A., Hassani, H., and Karbasi, A.
\newblock Conditional gradient method for stochastic submodular maximization:
  Closing the gap.
\newblock In \emph{AISTATS}, pp.\  1886--1895, 2018{\natexlab{a}}.

\bibitem[Mokhtari et~al.(2018{\natexlab{b}})Mokhtari, Hassani, and
  Karbasi]{mokhtari2018stochastic}
Mokhtari, A., Hassani, H., and Karbasi, A.
\newblock Stochastic conditional gradient methods: From convex minimization to
  submodular maximization.
\newblock \emph{arXiv preprint arXiv:1804.09554}, 2018{\natexlab{b}}.

\bibitem[Shalev-Shwartz et~al.(2012)]{shalev2012online}
Shalev-Shwartz, S. et~al.
\newblock Online learning and online convex optimization.
\newblock \emph{Foundations and Trends{\textregistered} in Machine Learning},
  4\penalty0 (2):\penalty0 107--194, 2012.

\bibitem[Streeter \& Golovin(2009)Streeter and Golovin]{streeter2009online}
Streeter, M. and Golovin, D.
\newblock An online algorithm for maximizing submodular functions.
\newblock In \emph{NIPS}, pp.\  1577--1584, 2009.

\bibitem[Vondr{\'a}k(2007)]{vondrak2007submodularity}
Vondr{\'a}k, J.
\newblock \emph{Submodularity in combinatorial optimization}.
\newblock PhD thesis, Charles University, 2007.

\bibitem[Vondr{\'a}k et~al.(2011)Vondr{\'a}k, Chekuri, and
  Zenklusen]{vondrak2011submodular}
Vondr{\'a}k, J., Chekuri, C., and Zenklusen, R.
\newblock Submodular function maximization via the multilinear relaxation and
  contention resolution schemes.
\newblock In \emph{STOC}, pp.\  783--792. ACM, 2011.

\bibitem[Wang \& Boyd(2008)Wang and Boyd]{Wang2008Fast}
Wang, Y. and Boyd, S.
\newblock Fast model predictive control using online optimization.
\newblock \emph{IFAC Proceedings Volumes}, 41\penalty0 (2):\penalty0 6974 --
  6979, 2008.
\newblock ISSN 1474-6670.
\newblock 17th IFAC World Congress.

\bibitem[Wolsey(1982)]{wolsey1982analysis}
Wolsey, L.~A.
\newblock An analysis of the greedy algorithm for the submodular set covering
  problem.
\newblock \emph{Combinatorica}, 2\penalty0 (4):\penalty0 385--393, 1982.

\bibitem[Xiao(2010)]{Xiao2010Dual}
Xiao, L.
\newblock Dual averaging methods for regularized stochastic learning and online
  optimization.
\newblock \emph{J. Mach. Learn. Res.}, 11:\penalty0 2543--2596, December 2010.
\newblock ISSN 1532-4435.

\bibitem[Yue \& Guestrin(2011)Yue and Guestrin]{Yue2011Linear}
Yue, Y. and Guestrin, C.
\newblock Linear submodular bandits and their application to diversified
  retrieval.
\newblock In \emph{NIPS}, pp.\  2483--2491, 2011.

\bibitem[Zachary(1977)]{zachary1977information}
Zachary, W.~W.
\newblock An information flow model for conflict and fission in small groups.
\newblock \emph{Journal of anthropological research}, 33\penalty0 (4):\penalty0
  452--473, 1977.

\bibitem[Zinkevich(2003)]{zinkevich2003online}
Zinkevich, M.
\newblock Online convex programming and generalized infinitesimal gradient
  ascent.
\newblock In \emph{ICML}, pp.\  928--936, 2003.

\end{thebibliography}
\bibliographystyle{icml2018}
\clearpage

\onecolumn
\appendix
\section{Variance Reduction Theorem}

Each of our results relies on a recent variance reduction technique, proposed by 
\cite{mokhtari2017conditional,mokhtari2018stochastic}. We now present 
Theorem~\ref{thm:variance_reduction}, which appears 
as Lemma~2 in~\cite{mokhtari2017conditional}. Although the proof is 
essentially the same, we present it here so that it is self-contained. When we apply 
Theorem~\ref{thm:variance_reduction} in the analysis of our algorithms, we will have that $ \{\ba_t \}$ 
are a sequence of 
gradients, $\{\bta_t\}$ are stochastic 
gradient estimates, and $\{\bd_t\}$ are the sequence of averaged gradient 
estimates. Moreover, the upper bound on the norm of the difference of gradients $ \| \ba_t - 
\ba_{t-1} \| $ comes from the iterate update procedure and smoothness of the objective function.

\begin{theorem}\label{thm:variance_reduction}
	 Let $ \{ 
	\ba_t\}_{t=0}^{T}$ be a sequence of points in $\reals^n$ 
	such that $ \| \ba_t - 
	\ba_{t-1} \| 
	\leq G/(t+s)  $ for all $1\leq t\leq T $ with fixed constants $ G\geq 0 
	$ and $ s\geq 3 $. 
	Let $ \{ \bta_t\}_{t=1}^T$ be a sequence of random variables such that $ \expect[ 
	\bta_t|\mathcal{F}_{t-1} ] = \ba_t $ and $ \expect[ \| \bta_t - \ba_t 
	\|^2|\mathcal{F}_{t-1} ] \leq \sigma^2$ for 
	every $ t\geq 0 $, 
	where 
	$ 
	\mathcal{F}_{t-1} $ is the $ \sigma $-field generated by 
	$ \{ \bta_i\}_{i=1}^{t} $ 
	and $ \mathcal{F}_{0} = \varnothing $. Let $\{\bd_t\}_{t=0}^T$ be a sequence of random 
	variables where $\bd_0$ is fixed and subsequent $\bd_{t}$ are obtained by the recurrence 
	$$ \bd_t = (1-\rho_t) \bd_{t-1} +\rho_t \bta_t $$ 
	with $ \rho_t = \frac{2}{(t+s)^{2/3}} $. 
	 Then, we have \[ 
	\expect[\| \ba_t-\bd_t\|^2 ] \leq \frac{Q}{(t+s+1)^{2/3}},
	\]
	where $ Q \triangleq \max \{ \|\ba_0 - \bd_0 \|^2 (s+1)^{2/3}, 
	4\sigma^2 + 3G^2/2 \} $.
\end{theorem}
We remark that we only need $ s\geq 2^{3/2} \approx 
2.83 $ in the statement of \cref{thm:variance_reduction}. 
\begin{proof}
	Let $ \Delta_t = \| \ba_t-\bd_t\|^2 $. We have the following identity
	\begin{dmath*}
		\Delta_t = \|  \rho_t (\ba_t-\bta_t)  + (1-\rho_t) (\ba_t-\ba_{t-1}) + 
		(1-\rho_t)(\ba_{t-1} - \bd_{t-1})  \|^2.
	\end{dmath*}
Expanding the square and taking the expectation with respect to $ 
\mathcal{F}_{t-1} $ gives
\begin{dmath*}
	\expect[\Delta_t | \mathcal{F}_{t-1}] \leq \rho_t^2 \sigma^2 + (1-\rho_t)^2 
	\frac{G^2}{(t+s)^2} + (1-\rho_t)^2  \Delta_{t-1} + 2 (1-\rho_t)^2
	\expect[ \langle 
	\ba_t-\ba_{t-1}, \ba_{t-1}-\bd_{t-1}\rangle|\mathcal{F}_{t-1} ].
\end{dmath*}
Taking the expectation again gives
\begin{dmath*}
	\expect[\Delta_t ] \leq \rho_t^2 \sigma^2 + (1-\rho_t)^2 
	\frac{G^2}{(t+s)^2} + (1-\rho_t)^2 \expect[ \Delta_{t-1}] + 2 
	(1-\rho_t)^2 \expect[ 
	\langle 
	\ba_t-\ba_{t-1}, \ba_{t-1}-\bd_{t-1}\rangle ].
\end{dmath*}
By Young's inequality, we have \begin{dmath*}
	2\langle 
	\ba_t-\ba_{t-1}, \ba_{t-1}-\bd_{t-1}\rangle \leq \beta_t 
	\|\ba_{t-1}-\bd_{t-1}\|^2 + (1/\beta_t)\frac{G^2}{(t+s)^2}.
\end{dmath*}
Therefore we deduce
\begin{dmath*}
	\expect[\Delta_t ] \leq \rho_t^2 \sigma^2 + (1-\rho_t)^2 
	\frac{G^2}{(t+s)^2} + (1-\rho_t)^2  \expect[\Delta_{t-1}] +  (1-\rho_t)^2\left( 
	\beta_t 
	\expect[\Delta_{t-1}] + (1/\beta_t)\frac{G^2}{(t+s)^2} \right)
	\leq \rho_t^2 \sigma^2 + 
	\frac{G^2}{(t+s)^2}(1-\rho_t)^2(1+\frac{1}{\beta_t}) + 
	\expect[\Delta_{t-1}](1-\rho_t)^2 (1+\beta_t).
\end{dmath*}
We write $ z_t $ for $ \expect[\Delta_t] $. Notice that $ 
(1-\rho_t)(1+\rho_t/2)\leq 1 $ as long as $ \rho_t\geq 0 $. If we assume $ 
\rho_t\in[0,1] $,
setting $ \beta_t=\rho_t/2 $ yields
\begin{dmath*}
	z_t \leq \rho_t^2 \sigma^2 + 
	\frac{G^2}{(t+s)^2}(1-\rho_t)^2(1+\frac{2}{\rho_t}) + 
	z_{t-1}(1-\rho_t)^2 (1+\frac{\rho_t}{2})
	\leq \rho_t^2 \sigma^2 + 
	\frac{G^2}{(t+s)^2}(1+\frac{2}{\rho_t}) + 
	z_{t-1}(1-\rho_t).
\end{dmath*}
We set $ \rho_t = \frac{2}{(t+s)^{2/3}} $, where $ s^{2/3}\geq 2 $. 
Since $ (t+s)^2 = (t+s)^{4/3}(t+s)^{2/3}\geq 2(t+s)^{4/3} $,
 we 
have
\begin{dmath*}
	z_t\leq (1-\frac{2}{(t+s)^{2/3}})z_{t-1} + \frac{4\sigma^2}{(t+s)^{4/3}} 
	+ \frac{G^2}{(t+s)^2} + \frac{G^2}{(t+s)^{4/3}}
	\leq (1-\frac{2}{(t+s)^{2/3}})z_{t-1} + \frac{4\sigma^2}{(t+s)^{4/3}} 
	 + \frac{3G^2}{2(t+s)^{4/3}}
	 \leq (1-\frac{2}{(t+s)^{2/3}})z_{t-1} + \frac{4\sigma^2 + 
	 3G^2/2}{(t+s)^{4/3}} 
 \leq (1-\frac{2}{(t+s)^{2/3}})z_{t-1} + \frac{Q}{(t+s)^{4/3}} .
\end{dmath*}
We claim $ z_t\leq \frac{Q}{(t+s+1)^{2/3}} $ for $ \forall 0\leq t\leq T $ and 
show this by induction. It holds for $ t=0 $ due to the definition of $ Q $. 
Now we assume that it is true for $ t=k-1 $. We have
\begin{dmath*}
	z_k\leq (1-\frac{2}{(k+s)^{2/3}})z_{k-1} + \frac{Q}{(k+s)^{4/3}}
	\leq (1-\frac{2}{(k+s)^{2/3}})\frac{Q}{(k+s)^{2/3}} + \frac{Q}{(k+s)^{4/3}}
	=  Q\frac{(k+s)^{2/3}-1}{(k+s)^{4/3}}.
\end{dmath*}
In order to show that $ z_k\leq \frac{Q}{(k+s+1)^{2/3}} $, it suffices to show 
that \[ 
((k+s)^{2/3}-1) (k+s+1)^{2/3} \leq (k+s)^{4/3}.
 \]
 The above inequality holds since $ (k+s+1)^{2/3} \leq (k+s)^{2/3}+1 $.
\end{proof}

\section{Proof of \cref{thm:adversarial}: Convex 
	Case}\label{app:adversarial_convex}


		We begin by examining the sequence of iterates $\bx_t^{(1)}, 
		\bx_t^{(2)}, 
		\dots,
		\bx_t^{(K+1)}$ produced in  
	\cref{alg:meta_frank_wolfe} for a fixed $t$. By 	
	definition of the update and because $f_t$ is $L$-smooth, 
	we have
	\begin{dmath*}
		f_t( \bx_t^{(k+1)}) - f_t(\bx^*) = f_t( \bx_t^{(k)} + \eta_{k} ( 
		\bv_t^{(k)} - \bx_t^{(k)}) ) - f_t(\bx^*) \\
		\leq f_t ( \bx_t^{(k)}) - f_t(\bx^*) + \eta_{k} \langle \nabla 
		f_t(\bx_t^{(k)}), \bv_t^{(k)} - \bx_t^{(k)} \rangle + \eta_{k}^2 
		\frac{L}{2} \| \bv_t^{(k)} - \bx_t^{(k)} \|^2 \\
		\leq f_t ( \bx_t^{(k)}) - f_t(\bx^*) + \eta_{k} \langle \nabla 
		f_t(\bx_t^{(k)}), \bv_t^{(k)} - \bx_t^{(k)} \rangle + \eta_{k}^2 
		\frac{LD^2}{2}. \\
	\end{dmath*}
	Now, observe that the dual pairing may be decomposed as
	\begin{dmath*}
		\langle \nabla f_t(\bx_t^{(k)}), \bv_t^{(k)} - \bx_t^{(k)} \rangle 
		= \langle \nabla f_t(\bx_t^{(k)}) - \bd_t^{(k)}, \bv_t^{(k)} - 
		\bx^* 
		\rangle 
		+ \langle \nabla f_t(\bx_t^{(k)}), \bx^* - \bx_t^{(k)}  \rangle 
		+ \langle \bd_t^{(k)}, \bv_t^{(k)} - \bx^* \rangle.
	\end{dmath*}
	 We can bound the first term using Young's Inequality to get 
	\begin{dmath*}
		\langle \nabla f_t(\bx_t^{(k)}) - \bd_t^{(k)}, \bv_t^{(k)} - \bx^* 
		\rangle  
		\leq \frac{1}{2 \beta_{k}} \| f_t(\bx_t^{(k)}) - \bd_t^{(k)} \|^2 
		+ 
		2 
		\beta_{k} \| \bv_t^{(k)} - \bx^* \|^2 \\
		\leq \frac{1}{2 \beta_{k}} \| f_t(\bx_t^{(k)}) - \bd_t^{(k)} \|^2 
		+ 
		2 
		\beta_{k} D^2
	\end{dmath*}
	for any $\beta_{k} > 0$, which will be chosen later in the proof. 
	We may also bound the second term in the decomposition of the dual 
	pairing using convexity of $f_t$, i.e. $\langle \nabla f_t(\bx_t^{(k)}), 
	 \bx^* - \bx_t^{(k)} \rangle \leq f_t( \bx^*) - f_t(\bx_t^{(k)})$. Using these 
	upper bounds, we get that
	\begin{dmath*}
		\langle \nabla f_t(\bx_t^{(k)}), \bv_t^{(k)} - \bx_t^{(k)} \rangle 
		\leq \frac{1}{2 \beta_{k}} \| f_t(\bx_t^{(k)}) - \bd_t^{(k)} \|^2 
		+ 
		2 
		\beta_{k} D^2 + f_t( \bx^*) - f_t(\bx_t^{(k)}) + \langle \bd_t^{(k)}, 
		\bv_t^{(k)} - \bx^* \rangle .
	\end{dmath*}
	Using this upper bound on the dual pairing in the first inequality, 
	we get that
	\begin{dmath*}
		f_t( \bx_t^{(k+1)}) - f_t(\bx^*) \leq (1 - \eta_{k} ) (f_t( 
		\bx_t^{(k)}) 
		- f_t(\bx^*)) + \eta_{k} \left[ \frac{1}{2 \beta_{k}} \| 
		f_t(\bx_t^{(k)}) - \bd_t^{(k)} \|^2 + 2 \beta_{k} D^2 +  \langle 
		\bd_t^{(k)}, \bv_t^{(k)} - \bx^* \rangle + \eta_{k} 
		\frac{LD^2}{2}\right].
	\end{dmath*}
	Now we will apply the variance reduction technique. Note that 
	\begin{dmath*}
		\| \nabla f_t(\bx_t^{(k+1)} - \nabla f_t ( \bx_t^{(k)}) \| \leq L \| \bx_t^{(k+1)} - \bx_t^{(k)} \| 
		\leq L \eta_k \| \bx_t^{(k)} - \bv_t^{(k)} \|
		\leq \frac{LD}{k + 3}
	\end{dmath*}
	Where we have used that $f_t$ is $L$-smooth, the convex update, and that the step size is $\eta_k 
	= \frac{1}{k+3}$. Now, using \cref{thm:variance_reduction} with $G=LD$ and $s=3$, we have that
	\begin{dmath*}
	\expect[\| f_t(\bx_t^{(k)}) - \bd_t^{(k)} \|^2] \leq 
	\frac{Q_t}{(k+4)^{2/3}} \leq \frac{Q}{(k+4)^{2/3}}
	\end{dmath*}. 
	Where $Q_t \triangleq \max \{ \|\nabla f_t(\bx_1) \|^2 4^{2/3}, 
	4\sigma^2 + 3(LD)^2/2 \}$ and 
	$Q \triangleq \max \{ 4^{2/3} \max_{1 \leq t \leq T} \|\nabla f_t(\bx_1) \|^2 , 
	4\sigma^2 + 3(LD)^2/2 \}$ 
	Thus, taking expectation of both sides of the optimality gap and 
	setting $\beta_{k} = \frac{Q^{1/2}}{2D(k+4)^{1/3}}$ yields
	\begin{dmath*}
		\expect[f_t( \bx_t^{(k+1)})] - f_t(\bx^*) 
		\leq 
		(1 - \eta_{k} ) (\expect[f_t( \bx_t^{(k)})] - f_t(\bx^*)) + 
		\eta_{k} \left[ \frac{2 Q^{1/2} D}{(k+4)^{1/3}} +  
		\langle \bd_t^{(k)}, \bv_t^{(k)} - \bx^* \rangle + 
		\eta_{k} \frac{LD^2}{2}\right].
	\end{dmath*}
	Now we have obtained an upper bound on the expected optimality gap 
	$\expect[f_t( \bx_t^{(k+1)})] - f_t(\bx^*)$ in terms 
	of the expected optimality gap $\expect[f_t( \bx_t^{(k)})] - f_t(\bx^*) $ 
	in the previous iteration. By induction on $k$, we get that the final iterate in the sequence,
	$\bx_t \triangleq \bx_t^{(K+1)}$, satisfies the following expected 
	optimality gap
	\begin{dmath} \label{eq:expected_op_gap_before_step_sizes}
		\expect[f_t(\bx_t)] - f_t(\bx^*) \leq 
		\prod_{k=1}^K(1 - \eta_k) \left[ f_t(\bx_1) - f_t(\bx^*) \right] + 
		\sum_{k=1}^K \eta_k \prod_{j=k+1}^{K} (1 - \eta_j ) 
		\left[  \frac{2 Q^{1/2} D}{(k+4)^{1/3}} +  \langle \bd_t^{(k)}, 
		\bv_t^{(k)} - 
		\bx^* \rangle + 
		\eta_k \frac{LD^2}{2} \right]
	\end{dmath}
	Recall that the Frank Wolfe step sizes are $\eta_k = \frac{1}{k+3}$. We may obtain upper bounds on 
	product of the form $\prod_{k=r}^K (1 - \eta_k) $ by 
	\begin{dmath*}
		{\prod_{k=r}^K (1 - \eta_k) 
		= \prod_{k=r}^K \left( 1 - \frac{1}{k+3} \right) 
		\leq \exp \left( - \sum_{k=r}^K 	\frac{1}{x+3} \right) 
		\leq \exp \left( - \int_{x=r}^{K+1} \frac{1}{x+3} dx \right)  
		=  \frac{r+3}{K+4} \leq \frac{r+3}{K}}
	\end{dmath*}
	Substituting step sizes $\eta_k = \frac{1}{k+3}$ into Eq 
	(\ref{eq:expected_op_gap_before_step_sizes}) and using this upper bound yields
	\begin{dmath} \label{eq:expected_op_gap_after_step_sizes}
		\expect[f_t(\bx_t)] - f_t(\bx^*) \leq 
		\frac{4}{K} \left[ f_t(\bx_1) - f_t(\bx^*) 
		\right] 
		+ 
		\sum_{k=1}^K \left( \frac{1}{k+3} \cdot \frac{k+4}{K} \right) 
		\left[  \frac{2 Q^{1/2} D}{(k+4)^{1/3}} +  \langle \bd_t^{(k)}, 
		\bv_t^{(k)} - 
		\bx^* \rangle + \frac{LD^2}{2 (k+3)} \right]
	\end{dmath}
	Which may be further simplified by using $\left( \frac{1}{k+3} \cdot \frac{k+4}{K} \right)  \leq 
	\frac{4}{3 K}$ to obtain 
	\begin{dmath*}
		\expect[f_t(\bx_t)] - f_t(\bx^*) \leq 
		\frac{4}{K} \left[ f_t(\bx_1) - f_t(\bx^*) \right] + 
		\frac{4}{3 K}
		\sum_{k=1}^K 
		\left[  \frac{2 Q^{1/2} D}{(k+3)^{1/3}} +  \langle \bd_t^{(k)}, 
		\bv_t^{(k)} - 
		\bx^* \rangle + \frac{LD^2}{2(k+3)} \right],
	\end{dmath*}
	As before, we can obtain the following upper bounds using integral methods:
	\begin{dmath*}
		{\sum_{k=1}^K \frac{1}{k+3} \leq \log \left( \frac{K+3}{3} \right) \leq \log(K+1)} 
		\quad \text{ and } \quad
		{\sum_{k=1}^K \frac{1}{(k+3)^{1/3}} \leq \frac{3}{2} \left( (K + 3)^{2/3} - 3^{2/3} \right) \leq 
		\frac{3}{2} K^{2/3}} 
	\end{dmath*}
	Substituting these bounds into Eq (\ref{eq:expected_op_gap_after_step_sizes}) yields
	\begin{dmath*}
		\expect[f_t(\bx_t)] - f_t(\bx^*) \leq 
		\frac{4}{K} \left[ f_t(\bx_1) - f_t(\bx^*) \right] + 
		\frac{4 Q^{1/2} D}{ K^{1/3}} + 
		\frac{4 LD^2 \log ( K + 1)}{3 K} +
		\frac{4}{3 K} \sum_{k=1}^K \langle \bd_t^{(k)}, \bv_t^{(k)} - \bx^* 
		\rangle .
	\end{dmath*}
	Now, we can begin to bound regret by summing over all $t=1 \dots T$ to obtain
	\begin{dmath*}
		\sum_{t=1}^T \expect[f_t(\bx_t)] - \sum_{t=1}^T f_t(\bx^*) 
		\leq \frac{4}{K} \sum_{t=1}^T \left[ f_t(\bx_1) - f_t(\bx^*) \right] +
		\frac{4 T Q^{1/2} D}{K^{1/3}} + 
		\frac{4 TLD^2 \log ( K + 1)}{3 K} +
		\frac{4}{3 K} \sum_{t=1}^T \sum_{k=1}^K \langle \bd_t^{(k)}, \bv_t^{(k)} 
		- 
		\bx^* \rangle 
	\end{dmath*}
	Recall that for a fixed $k$, the sequence $\{ \bv_t^{(k)} \}_{t=1}^T$ is 
	produced by a online linear minimization oracle with regret $\oregret$ so 
	that
	\begin{dmath*}
		{\sum_{t=1}^T \langle \bd_t^{(k)}, \bv_t^{(k)} - \bx^* \rangle 
			\leq \sum_{t=1}^T \langle \bd_t^{(k)}, \bv_t^{(k)} \rangle - 
			\min_{\bx \in 
				\constraint} \sum_{t=1}^T \langle \bd_t^{(k)}, \bx \rangle \leq 
			\oregret}.
	\end{dmath*}
	Substituting this into the upper bound and using $M= \max_{1\leq t \leq T} \left[ f_t(\bx_1) - 
	f_t(\bx^*) \right]$ yields
	\begin{dmath*}
		\sum_{t=1}^T \expect[f_t(\bx_t)] - \sum_{t=1}^T f_t(\bx^*) 
		\leq
		\frac{4 T D Q^{1/2}}{K^{1/3}} + \frac{4 T}{K} \left(M + \frac{LD^2}{3} \log (K +1)
		\right) + \frac{4}{3} \oregret
	\end{dmath*}
	Now, setting $K = T^{3/2}$ and using a linear oracle with $\oregret = O(\sqrt{T})$ yields
	\begin{dmath*}
		\sum_{t=1}^T \expect[f_t(\bx_t)] - \sum_{t=1}^T f_t(\bx^*) 
		\leq 
		4 \sqrt{T} D Q^{1/2} + \frac{4}{\sqrt{T}} \left(M + \frac{LD^2}{3} 
		(\log T^{3/2} +1)
		\right) + \frac{4}{3} \oregret
		= O(\sqrt{T}).
	\end{dmath*}

\section{Proof of \cref{thm:adversarial}: DR-Submodular 
Case}\label{app:adversarial_submodular}
 Using the smoothness of $f_t$ and recalling $\bx_t^{(k+1)}-\bx_t^{(k)} = 
 \frac{1}{K}\bv_t^{(k)}$, we have
 \begin{dmath}\label{eq:using}
f_t(\bx_t^{(k+1)}) \geq f_t(\bx_t^{(k)}) + \langle \nabla 
f_t(\bx_t^{(k)}),\bx_t^{(k+1)}-\bx_t^{(k)}\rangle -\frac{L}{2} \lVert 
\bx_t^{(k+1)}-\bx_t^{(k)}\rVert^2
= f_t(\bx_t^{(k)}) + \langle \frac{1}{K} \nabla 
f_t(\bx_t^{(k)}),\bv_t^{(k)}\rangle -\frac{L}{2K^2} \lVert 
\bv_t^{(k)}\rVert^2
\geq f_t(\bx_t^{(k)}) + \frac{1}{K} \langle  \nabla 
f_t(\bx_t^{(k)}),\bv_t^{(k)}\rangle -\frac{LD^2}{2K^2}.
 \end{dmath}
We can re-write the term $\langle  \nabla 
f_t(\bx_t^{(k)}),\bv_t^{(k)}\rangle$ as 
\begin{dmath}\label{eq:re-write}
\langle  \nabla f_t(\bx_t^{(k)}),\bv_t^{(k)}\rangle = \langle  \nabla 
f_t(\bx_t^{(k)}) - \bd_t^{(k)},\bv_t^{(k)}\rangle + \langle   
\bd_t^{(k)},\bv_t^{(k)}\rangle
= \langle  \nabla f_t(\bx_t^{(k)}) - \bd_t^{(k)},\bv_t^{(k)}\rangle + 
\langle   \bd_t^{(k)}, \bx^*\rangle+
\langle   \bd_t^{(k)},\bv_t^{(k)} - \bx^*\rangle
= \langle  \nabla f_t(\bx_t^{(k)}) - 
\bd_t^{(k)},\bv_t^{(k)}-\bx^*\rangle + 
\langle   \nabla f_t(\bx_t^{(k)}), \bx^*\rangle+
\langle   \bd_t^{(k)},\bv_t^{(k)} - \bx^*\rangle.
\end{dmath}
We claim $\langle \nabla f_t(\bx_t^{(k)}), \bx^*\rangle \geq f_t(\bx^*) 
- f_t(\bx_t^{(k)})$. Indeed, using monotonicity of $f_t$ and 
 concavity along non-negative directions, we have
\begin{dmath}\label{eq:in-fact}
f_t(\bx^*)-f_t(\bx_t^{(k)}) \leq f_t(\bx^* \vee \bx_t^{(k)} )-f_t(\bx_t^{(k)}) 
\leq \langle \nabla f_t(\bx_t^{(k)}), \bx^* \vee \bx_t^{(k)} - 
\bx_t^{(k)}\rangle
= \langle \nabla f_t(\bx_t^{(k)}), (\bx^*- \bx_t^{(k)}) \vee 0 \rangle
\leq \langle \nabla f_t(\bx_t^{(k)}), \bx^* \rangle.
\end{dmath}
Plugging \cref{eq:in-fact} into \cref{eq:re-write}, we obtain 
\begin{dmath}\label{eq:obtain}
\langle  \nabla f_t(\bx_t^{(k)}),\bv_t^{(k)}\rangle \geq \langle  \nabla 
f_t(\bx_t^{(k)}) - \bd_t^{(k)},\bv_t^{(k)}-\bx^*\rangle + 
\langle   \bd_t^{(k)},\bv_t^{(k)} - \bx^*\rangle +
(f_t(\bx^*)-f_t(\bx_t^{(k)})).
\end{dmath}
Using Young's inequality, we can show that 
\begin{dmath}\label{eq:young}
\langle  \nabla f_t(\bx_t^{(k)}) - \bd_t^{(k)},\bv_t^{(k)}-\bx^*\rangle 
\geq -\frac{1}{2\beta^{(k)}} \lVert \nabla f_t(\bx_t^{(k)}) - 
\bd_t^{(k)} \rVert^2 - \frac{\beta^{(k)}}{2} \lVert  
\bv_t^{(k)}-\bx^* \rVert^2
\geq -\frac{1}{2\beta^{(k)}} \lVert \nabla f_t(\bx_t^{(k)}) - 
\bd_t^{(k)} \rVert^2 - \beta^{(k)} D^2/2
\end{dmath}
Then we plug \cref{eq:obtain,eq:young} into \cref{eq:using}, we deduce
\begin{dmath*}
f_t(\bx_t^{(k+1)}) \geq 
f_t(\bx_t^{(k)}) + \frac{1}{K} \left[-\frac{1}{2\beta^{(k)}} 
\lVert\nabla f_t(\bx_t^{(k)}) - \bd_t^{(k)} \rVert^2 - \beta^{(k)} D^2/2 
+ 
\langle   \bd_t^{(k)},\bv_t^{(k)} - \bx^*\rangle +
(f_t(\bx^*)-f_t(\bx_t^{(k)}))\right] -\frac{LD^2}{2K^2}.
\end{dmath*}
Equivalently, we have
\begin{dmath}\label{eq:recursive}
f_t(\bx^*) - f_t(\bx_t^{(k+1)}) \leq (1-1/K)[f_t(\bx^*) - 
f_t(\bx_t^{(k)})] - \frac{1}{K} \left[-\frac{1}{2\beta^{(k)}} \lVert 
\nabla f_t(\bx_t^{(k)}) - \bd_t^{(k)} \rVert^2 - \beta^{(k)} D^2/2 + 
\langle   \bd_t^{(k)},\bv_t^{(k)} - \bx^*\rangle 
\right] + \frac{LD^2}{2K^2}.
\end{dmath}
Applying \cref{eq:recursive} recursively for $1\leq k\leq K$ 
immediately yields
\begin{dmath*}
f_t(\bx^*) - f_t(\bx_t^{(k+1)}) \leq (1-1/K)^K [f_t(\bx^*) - 
f_t(\bx_t^{(1)})] + \frac{1}{K} \sum_{k=1}^{K} 
\left[\frac{1}{2\beta^{(k)}} \lVert \nabla f_t(\bx_t^{(k)}) - 
\bd_t^{(k)} \rVert^2 + \beta^{(k)} D^2/2 + 
\langle   \bd_t^{(k)}, \bx^*- \bv_t^{(k)} \rangle 
\right] + \frac{LD^2}{2K}.
\end{dmath*}
Recall that the point played in round $t$ is $ \bx_t \triangleq \bx_t^{(K+1)} $, the first 
iterate in the sequence is $\bx_t^{(1)} = 0$, and 
that
$(1-1/K)^K\leq 1/e$ for all $K \geq 1$ so that
\begin{dmath*}
f_t(\bx^*) - f_t(\bx_t) \leq \frac{1}{e}[f_t(\bx^*) - 
f_t(0)] + \frac{1}{K} \sum_{k=1}^{K} \left[\frac{1}{2\beta^{(k)}} 
\lVert \nabla f_t(\bx_t^{(k)}) - \bd_t^{(k)} \rVert^2 + \beta^{(k)} D^2/2 
+ 
\langle   \bd_t^{(k)}, \bx^*- \bv_t^{(k)} \rangle 
\right] + \frac{LD^2}{2K}.
\end{dmath*}
Since $f_t(0)\geq 0$, we obtain
\begin{dmath}\label{eq:should-sum}
(1-1/e)f_t(\bx^*) - f_t(\bx_t) \leq  \frac{1}{K} \sum_{k=1}^{K} 
\left[\frac{1}{2\beta^{(k)}} \lVert \nabla f_t(\bx_t^{(k)}) - 
\bd_t^{(k)} \rVert^2 + \beta^{(k)} D^2/2 + 
\langle   \bd_t^{(k)}, \bx^*- \bv_t^{(k)} \rangle 
\right] + \frac{LD^2}{2K}.
\end{dmath}
If we sum \cref{eq:should-sum} over $t=1,2,3,\ldots,T$, we obtain
\begin{dmath*}
(1-1/e) \sum_{t=1}^T f_t(\bx^*) - \sum_{t=1}^T f_t(\bx_t) \leq  
\frac{1}{K} \sum_{k=1}^{K} \left[\frac{1}{2\beta^{(k)}} 
\sum_{t=1}^T\lVert \nabla f_t(\bx_t^{(k)}) - \bd_t^{(k)} \rVert^2 + 
\beta^{(k)} D^2 T/2 + 
\sum_{t=1}^T \langle   \bd_t^{(k)}, \bx^*- \bv_t^{(k)} \rangle 
\right] + \frac{LD^2T}{2K}.
\end{dmath*}
By the definition of the regret, we have
\begin{dmath*}
\sum_{t=1}^T \langle   \bd_t^{(k)}, \bx^*- \bv_t^{(k)} \rangle \leq 
\oregret.
\end{dmath*}
Therefore, we deduce 
\begin{dmath*}
(1-1/e) \sum_{t=1}^T f_t(\bx^*) - \sum_{t=1}^T f_t(\bx_t)\\ \leq  
\frac{1}{K} \sum_{k=1}^{K} \left[\frac{1}{2\beta^{(k)}} 
\sum_{t=1}^T\lVert \nabla f_t(\bx_t^{(k)}) - \bd_t^{(k)} \rVert^2 + 
\beta^{(k)} D^2 T/2
\right] + \frac{LD^2T}{2K} + \oregret.
\end{dmath*}
Taking the expectation in both sides, we obtain
\begin{dmath}\label{eq:expect}
(1-1/e) \sum_{t=1}^T \expect[f_t(\bx^*)] - \sum_{t=1}^T 
\expect[f_t(\bx_t)]\\ \leq  \frac{1}{K} \sum_{k=1}^{K} 
\left[\frac{1}{2\beta^{(k)}} \sum_{t=1}^T \expect[ \lVert \nabla 
f_t(\bx_t^{(k)}) - \bd_t^{(k)} \rVert^2] + \beta^{(k)} D^2 T/2
\right] + \frac{LD^2T}{2K} + \oregret.
\end{dmath}
Notice that $ \lVert \nabla 
f_t(\bx_t^{(k)}) - \nabla f_t(\bx_t^{(k-1)}) \rVert \leq L\| \bv_t^{(k)} 
\|/T \leq LR/T\leq 2LR/(k+3) $.
By \cref{thm:variance_reduction}, if we set $\rho_k = 
\frac{2}{(k+3)^{2/3}}$, we have
\begin{dmath}\label{eq:Q}
\expect[ \lVert \nabla f_t(\bx_t^{(k)}) - \bd_t^{(k)} \rVert^2] \leq 
\frac{Q_t}{(k+4)^{2/3}} \leq \frac{Q}{(k+4)^{2/3}},
\end{dmath}
where $Q_t\triangleq \max\{ \lVert \nabla f_t(0) \rVert^2 4^{2/3}, 
 4\sigma^2 + 6L^2 R^2 \}$ and $Q \triangleq \max\{ \max_{1\leq t\leq 
	T} \lVert \nabla f_t(\bx_1) 
\rVert^2 4^{2/3}, 4\sigma^2 + 6L^2 R^2 \}$.

Plugging \cref{eq:Q} into \cref{eq:expect} and setting $\beta^{(k)} = 
(Q^{1/2})/ (D(k+3)^{1/3})$, we deduce
\begin{dmath*}
(1-1/e) \sum_{t=1}^T \expect[f_t(\bx^*)] - \sum_{t=1}^T 
\expect[f_t(\bx_t)] \leq  \frac{TDQ^{1/2}}{K} \sum_{k=1}^{K} 
\frac{1}{(k+4)^{1/3}} + \frac{LD^2T}{2K} + \oregret
\end{dmath*}
Since $\sum_{k=1}^{K} \frac{1}{(k+4)^{1/3}} \leq \int_0^K 
\frac{dx}{(x+4)^{1/3}} = \frac{3}{2}[(K+4)^{2/3}-9^{2/3}]\leq 
\frac{3}{2}K^{2/3}$, we have
\begin{dmath*}
(1-1/e) \sum_{t=1}^T \expect[f_t(\bx^*)] - \sum_{t=1}^T 
\expect[f_t(\bx_t)] \leq  \frac{3TDQ^{1/2}}{2K^{1/3}}  + 
\frac{LD^2T}{2K} + \oregret.
\end{dmath*}

\section{Proof of \cref{thm:stochastic}: Convex 
	Case}\label{app:stochastic_convex}
Let $f(\bx) = \expect_{f_t \sim \mathcal{D}} [f_t(\bx)]$ denote the expected function. Because $f$ is 
$L$-smooth and 
convex, we have
\begin{dmath*}
	f( \bx_{t+1}) - f(\bx^*) = f( \bx_t + \eta_t ( 
	\bv_t - \bx_t) ) - f(\bx^*) \\
	\leq f ( \bx_t) - f(\bx^*) + \eta_t \langle \nabla 
	f(\bx_t), \bv_t - \bx_t \rangle + \eta_{t}^2 
	\frac{L}{2} \| \bv_t - \bx_t \|^2 \\
	\leq f ( \bx_t) - f_t(\bx^*) + \eta_t \langle \nabla 
	f(\bx_t), \bv_t - \bx_t \rangle + \eta_t^2 
	\frac{LD^2}{2}. \\
\end{dmath*}
As before, the dual pairing may be decomposed as 
\begin{dmath*}
	\langle \nabla f(\bx_t), \bv_t - \bx_t \rangle = \langle \nabla f(\bx_t) - 
	\bd_t, \bv_t - \bx^* \rangle + 
	\langle \nabla f(\bx_t), \bx^* - \bx_t \rangle + \langle \bd_t, \bv_t - 
	\bx^* \rangle.
\end{dmath*}
We can bound the first term using Young's Inequality to get 
\begin{dmath*}
	\langle \nabla f(\bx_t) - \bd_t, \bv_t - \bx^* 
	\rangle  
	\leq \frac{1}{2 \beta} \| f(\bx_t) - \bd_t \|^2 + 2 
	\beta \| \bv_t - \bx^* \|^2 \\
	\leq \frac{1}{2 \beta} \| f(\bx_t) - \bd_t \|^2 + 2 
	\beta D^2.
\end{dmath*}
for any $\beta > 0$, which will be chosen later in the proof. 
We may also bound the second term in the decomposition of the dual 
pairing using convexity of $f$, i.e. $\langle \nabla f(\bx_t), 
\bx^* - \bx_t \rangle \leq f_t( \bx^*) - f(\bx_t)$. Finally, the third term is 
nonpositive, by the choice of 
$\bv_t$, namely $\bv_t = \argmin_{\bv \in \constraint} \langle \bd_t, \bv 
\rangle$. Using these 
inequalities, we now have that
\begin{dmath*} 
	f( \bx_{t+1}) - f(\bx^*)  
	\leq 
	( 1 - \eta_t) \left( f(\bx_t) - f(\bx^*) \right) + 
	\eta_t \left( \frac{1}{2 \beta} \| f(\bx_t) - \bd_t \|^2 + 2 \beta D^2 
	\right) + 
	\eta_t^2 \frac{LD^2}{2}.
\end{dmath*}
Taking expectation over the randomness in the iterates (i.e. the stochastic 
gradient estimates), we 
have that
\begin{dmath} \label{eq:opt_gap_before_vr}
	\expect[f( \bx_{t+1})] - f(\bx^*)  
	\leq 
	( 1 - \eta_t) \left( \expect[f(\bx_t)] - f(\bx^*) \right) + 
	\eta_t \left( \frac{1}{2 \beta} \expect[\| f(\bx_t) - \bd_t \|^2] + 2 \beta 
	D^2 \right) + 
	\eta_t^2 \frac{LD^2}{2}.
\end{dmath}
Now we will apply the variance reduction technique. Note that 
\begin{dmath*}
	\| \nabla f(\bx_{t+1}) - \nabla f ( \bx_t) \| \leq L \| \bx_{t+1} - \bx_t \| 
	\leq L \eta_t \| \bx_t - \bv_t\|
	\leq  L \eta_t D
\end{dmath*}
where we have used that $f$ is $L$-smooth, the convex update, and the diameter. Now, using 
\cref{thm:variance_reduction} with $G=LD$ and $s=3$, we have that
\begin{dmath*}
	\expect[\| f(\bx_t) - \bd_t \|^2] \leq  \frac{Q}{(t+4)^{2/3}},
\end{dmath*}
where 
$Q \triangleq \max \{ 4^{2/3} \|\nabla f(\bx_1) \|^2 , 
4\sigma^2 + 3(LD)^2/2 \}$.
 Using this bound in Eq (\ref{eq:opt_gap_before_vr}) and setting $\beta = 
\frac{Q^{1/2}}{2D(t+4)^{1/3}}$ yields
\begin{dmath*}
	\expect[f( \bx_{t+1})] - f(\bx^*)  
	\leq 
	\left(1 - \eta_t \right) \left( \expect[f(\bx_t)] - f(\bx^*) \right) + 
	\eta_t \frac{2 Q^{1/2} D}{(t+4)^{1/3}} + 
	\eta_t^2 \frac{LD^2}{2}.
\end{dmath*}
By induction, we have
\begin{dmath*}
	\expect[f( \bx_{t+1})] - f(\bx^*)  
	\leq 	
	\prod_{k=1}^t \left( 1 -\eta_k \right)  M  +
	\sum_{k=1}^t \eta_k \prod_{j=k+1}^t \left( 1 - \eta_j \right) \left( 
	\frac{2 Q^{1/2} 
	D}{(k+4)^{1/3}} + \eta_k
	\frac{LD^2}{2} \right),
\end{dmath*}
where $ M=  f(\bx_1) - f(\bx^*)  $.
Recall that the step size is set to be $\eta_t = \frac{1}{t+3}$.
As in Appendix~\ref{app:adversarial_convex}, we can obtain the bounds 
$\prod_{k=1}^t (1 - \eta_k) = \prod_{k=1}^t (1 - \frac{1}{k+3}) \leq 
\exp(-\sum_{k=1}^{t} \frac{1}{k+3})\leq \exp(-\int_1^{t+1}\frac{dx}{x+3} ) = 4/(t+4) $ and 
similarly $ \prod_{j=k+1}^t (1 - \frac{1}{j+3}) \leq \frac{k+4}{t+4} $. Using these bounds as well 
as the choice of step size $\eta_t = \frac{1}{t+3}$ in the above yields
\begin{dmath*}
	\expect[f( \bx_{t+1})] - f(\bx^*)  
	\leq 	
	\frac{4 M}{t+4}  +
	\sum_{k=1}^t \left( \frac{1}{k+3} \cdot \frac{k+4}{t+4} \right) 
	\left( 
	\frac{2 Q^{1/2} D}{(k+4)^{1/3}} 
	+ \frac{1}{k+3} \frac{LD^2}{2} 
	\right)
	= 
	\frac{4 M}{t+4}  +
	\frac{4}{3 (t+4)} \sum_{k=1}^t
	\left( 
	\frac{2 Q^{1/2} D}{(k+4)^{1/3}} 
	+ \frac{1}{k+3} \frac{LD^2}{2} 
	\right)
\end{dmath*}
where the second inequality used $\left( \frac{1}{k+3} \cdot \frac{k+4}{(t+4)} \right) < \frac{4}{3(t+4)}$.
As before in Section~\ref{app:adversarial_convex}, using the inequalities 
$\sum_{k=1}^t \frac{1}{k+3} \leq \log(t+1)$ 
and 
$\sum_{k=1}^t \frac{1}{(k+3)^{1/3}} \leq \frac{3}{2} t^{2/3}$ in the above yields
\begin{equation}\label{eq:single-regret}
	\expect[f( \bx_{t+1})] - f(\bx^*)  
	\leq \frac{4 M }{t+4}  + 
	4 Q^{1/2}D \frac{t^{2/3}}{t+4}  + 
	\frac{4}{3} LD^2 \frac{\log(t+1)}{t+4}.
\end{equation}
To obtain a regret bound, we sum over rounds $t=1, \dots T$ to obtain
\begin{equation*}
\sum_{t=1}^T \expect[f(\bx_t)] - T f(\bx^*) 
	\leq 4 M \left( \sum_{t=1}^T \frac{1}{t+4} \right)
	+ 4 Q^{1/2}D \left( \sum_{t=1}^T \frac{t^{2/3}}{t+4} \right) 
	+ \frac{4}{3} LD^2  \left( \sum_{t=1}^T \frac{\log(t+1)}{t+4} \right)
\end{equation*}
Using the integral trick again, we obtain the upper bounds $ 
\sum_{t=1}^{T}\frac{1}{t+4} \leq \log(T+1) $, 
$ \sum_{t=1}^{T}\frac{t^{2/3}}{t+4} \leq \frac{3}{2}T^{2/3} $, 
and 
$ \sum_{t=1}^{T} \frac{\log(t+3)}{t+4} \leq \log^2(T+3) $. Substituting these bounds in the regret bound 
above yields 
\begin{equation*}
\sum_{t=1}^T \expect[f(\bx_t)] - T f(\bx^*) 
\leq 4 M \log(T+1)
+ 6 Q^{1/2}D T^{2/3} 
+ \frac{4}{3} LD^2 \log^2(T+3) = O\left( T^{2/3} \right)
\end{equation*}

\section{Proof of \cref{thm:stochastic}: DR-Submodular 
Case}\label{app:stochastic_submodular}

Since $f$ is $L$-smooth, we obtain 
\begin{dmath*}
f(\bx_{t+1})\geq f(\bx_t) + \langle \nabla f(\bx_t), \frac{1}{T} 
\bv_t \rangle - \frac{L}{2} \lVert \frac{1}{T} \bv_t \rVert^2
\geq f(\bx_t) + \frac{1}{T} \langle \nabla f(\bx_t),  \bv_t \rangle - 
\frac{LD^2}{2T^2} 
= f(\bx_t) + \frac{1}{T} \langle \bd_t,  \bv_t \rangle + \frac{1}{T} 
\langle \nabla f(\bx_t) - \bd_t,  \bv_t \rangle - \frac{LD^2}{2T^2} 
\geq f(\bx_t) + \frac{1}{T} \langle \bd_t,  \bx^* \rangle + 
\frac{1}{T} \langle \nabla f(\bx_t) - \bd_t,  \bv_t \rangle - 
\frac{LD^2}{2T^2} 
= f(\bx_t) + \frac{1}{T} \langle \nabla f(\bx_t)- \bd_t, \bv_t - 
\bx^* \rangle + \frac{1}{T} \langle f(\bx_t), \bx^*\rangle - 
\frac{LD^2}{2T^2}. 
\end{dmath*}
In the last inequality, we used the fact that $\bv_t = \argmax_{\bv 
\in \constraint} \langle \bd_t, \bv\rangle$. Similar to 
\cref{eq:in-fact} in \cref{app:adversarial_submodular}, we have $\langle 
f(\bx_t), \bx^*\rangle \geq f(\bx^*) - f(\bx_t)$. Again, Young's 
inequality gives $\langle \nabla f(\bx_t)- \bd_t, \bv_t - \bx^* 
\rangle \geq -\frac{1}{2}(\beta_t \lVert \bv_t - \bx^* \rVert^2 + 
\lVert f(\bx_t)- \bd_t \rVert^2 / \beta_t )$. Therefore, we deduce
\begin{dmath*}
f(\bx_{t+1}) \geq f(\bx_t) - \frac{1}{2T}(\beta_t \lVert \bv_t - 
\bx^* \rVert^2 + \lVert f(\bx_t)- \bd_t \rVert^2 / \beta_t) + 
\frac{1}{T} (f(\bx^*) - f(\bx_t)) - \frac{LD^2}{2T^2}
\geq f(\bx_t) - \frac{1}{2T}(\beta_t D^2  + \lVert f(\bx_t)- \bd_t 
\rVert^2 / \beta_t) + \frac{1}{T} (f(\bx^*) - f(\bx_t)) - 
\frac{LD^2}{2T^2}.
\end{dmath*}
Re-arrangement of the terms yields 
\begin{dmath*}
f(\bx^*)-f(\bx_{t+1})\leq (1-1/T)(f(\bx^*)-f(\bx_t)) + 
\frac{1}{2T}(\beta_t D^2  + \lVert f(\bx_t)- \bd_t \rVert^2 / 
\beta_t) + \frac{LD^2}{2T^2}.
\end{dmath*}
Recalling that $(1-1/T)^T\leq 1/e$ and $f(\bx_1)=f(0)\geq 0$, we have
\begin{dmath*}
f(\bx^*)-f(\bx_{t+1}) \leq (1-1/T)^t 
(f(\bx^*)-f(\bx_1))+\frac{1}{2T}\sum_{i=1}^t (\beta_i D^2  + \lVert 
f(\bx_i)- \bd_i \rVert^2 / \beta_i) + \frac{LD^2}{2T}
\leq \frac{1}{e} f(\bx^*)+\frac{1}{2T}\sum_{i=1}^t (\beta_i D^2  + 
\lVert f(\bx_i)- \bd_i \rVert^2 / \beta_i) + \frac{LD^2}{2T}, 
\end{dmath*}
which in turn yields
\begin{dmath*}
(1-1/e)f(\bx^*)-f(\bx_{t+1}) \leq \frac{1}{2T}\sum_{i=1}^t (\beta_i 
D^2  + \lVert f(\bx_i)- \bd_i \rVert^2 / \beta_i) + \frac{LD^2}{2T}.
\end{dmath*}
Taking expectation in both sides gives
\begin{dmath*}
(1-1/e)\expect[f(\bx^*)]-\expect[f(\bx_{t+1})] \leq 
\frac{1}{2T}\sum_{i=1}^t (\beta_i D^2  + \expect[\lVert f(\bx_i)- 
\bd_i \rVert^2] / \beta_i) + \frac{LD^2}{2T}.
\end{dmath*}
Notice that $ \lVert \nabla 
f(\bx_{t}) - \nabla f(\bx_{t-1}) \rVert \leq L\| \bv_t 
\|/K \leq LR/K\leq 2LR/(k+3) $.
By \cref{thm:variance_reduction}, if we set $\rho_i = 
\frac{2}{(i+3)^{2/3}}$, we have 
\begin{dmath*}
\expect[\lVert f(\bx_i)- \bd_i \rVert^2] \leq \frac{Q}{(i+4)^{2/3}}
\end{dmath*}
for every $i\leq T$ and $Q =\max\{ 
\lVert \nabla f(0) 
\rVert^2 4^{2/3}, 4\sigma^2 + 6L^2 R^2 \} $. 
If we set $\beta_i = \frac{Q^{1/2}}{D(i+4)^{1/3}}$, we have
\begin{dmath*}
(1-1/e)\expect[f(\bx^*)]-\expect[f(\bx_{t+1})] \leq \sum_{i=1}^t 
\frac{DQ^{1/2}}{(i+4)^{1/3}T} + \frac{LD^2}{2T} 
\leq \frac{3DQ^{1/2}t^{2/3}}{2T} + \frac{LD^2}{2T} 
\end{dmath*}
since $\sum_{i=1}^t \frac{1}{(i+4)^{1/3}} \leq \int_0^t 
\frac{1}{(x+4)^{1/3}} dx = \frac{3}{2}[(x+4)^{2/3}]^t_0 \leq 
\frac{3}{2}[x^{2/3}]^{t}_{0} = \frac{3}{2}t^{2/3}$.

Therefore we have
\begin{dmath*}
(1-1/e)T \expect[f(\bx^*)]-  \sum_{t=1}^T \expect[f(\bx_{t})] \\
=  (1-1/e)\expect[f(\bx^*)]-f(0) + \sum_{t=1}^{T-1} 
[(1-1/e)\expect[f(\bx^*)]-\expect[f(\bx_t)]] \\
\leq (1-1/e)\expect[f(\bx^*)]-f(0) + \sum_{t=1}^{T-1} \left[ 
\frac{3DQ^{1/2}t^{2/3}}{2T} + \frac{LD^2}{2T}  \right] .
\end{dmath*}
Since $\sum_{t=1}^{T-1} t^{2/3} = 1 + \sum_{t=2}^{T-1} t^{2/3} \leq 
1+ \int_1^{T} t^{2/3} dt = \frac{3}{5}T^{5/3}+\frac{2}{5} $, we 
conclude
\begin{dmath*}
(1-1/e)T\expect[f(\bx^*)]-  \sum_{t=1}^T \expect[f(\bx_{t})]
\leq (1-1/e)\expect[f(\bx^*)]-f(0) + 
\frac{3DQ^{1/2}}{10}(3T^{2/3}+2T^{-1})+\frac{LD^2}{2} = O(T^{2/3}).
\end{dmath*}

\end{document}